\documentclass[10pt,twocolumn,letterpaper]{article}
\usepackage{cvpr}
\usepackage{times}
\usepackage{epsfig}
\usepackage{graphicx}
\usepackage{amsmath}
\usepackage{amssymb}
\usepackage{multirow}
\usepackage{caption}
\usepackage{subcaption}
\usepackage{comment}

\usepackage[pagebackref=true,breaklinks=true,letterpaper=true,colorlinks,bookmarks=false]{hyperref}

 \cvprfinalcopy 

\def\cvprPaperID{3266} 

\ifcvprfinal\fi
\begin{document}

\title{Self-critical Sequence Training for Image Captioning}

\author{Steven J. Rennie$^{1}$, Etienne Marcheret$^{1}$, Youssef Mroueh, Jerret Ross and Vaibhava Goel$^{1}$\\
Watson Multimodal Algorithms and Engines Group \\
IBM T.J. Watson Research Center, NY, USA \\
{\hspace{-0.20in}\tt\small{steve@fusemachines.com, \{etiennemarcheret, vaibhavagoel\}@gmail.com}, \{mroueh, rossja\}@us.ibm.com}
}

\maketitle

{\let\thefootnote\relax\footnotetext{$^{1}$Authors Steven J. Rennie, Etienne Marcheret, and Vaibhava Goel were at IBM while the work was being completed.}} 
\begin{abstract}
Recently it has been shown that policy-gradient methods for reinforcement learning can be utilized to
train deep end-to-end systems directly on non-differentiable metrics for the task at hand. In this paper we consider the problem of optimizing image captioning systems using reinforcement learning, and show that by carefully optimizing our systems using the test metrics of the MSCOCO task, significant gains in performance can be realized. Our systems are built using a new optimization approach that we call self-critical sequence training (SCST). SCST is a form of the popular REINFORCE algorithm that, rather than estimating a ``baseline" to normalize the rewards and reduce variance, utilizes the output of its own test-time inference algorithm to normalize the rewards it experiences. Using this approach, estimating the reward signal (as actor-critic methods must do) and estimating normalization (as REINFORCE algorithms typically do) is avoided, while at the same time harmonizing the model with respect to its test-time inference procedure. Empirically we find that directly optimizing the CIDEr metric with SCST and greedy decoding at test-time is highly effective.  Our results on the MSCOCO evaluation sever establish a new state-of-the-art on the task, improving the best result in terms of CIDEr from 104.9 to 114.7.
\end{abstract}

\section{Introduction}
Image captioning aims at generating a natural language description of an image. Open domain captioning is a very challenging task, as it requires a fine-grained understanding of the global and the local entities in an image, as well as their attributes and relationships. 
The recently released MSCOCO challenge \cite{MSCOCO} provides a new, larger scale platform for evaluating image captioning systems, complete with an evaluation server for benchmarking competing methods. Deep learning approaches to sequence modeling have yielded impressive results on the task, dominating the task leaderboard. Inspired by the recently introduced encoder/decoder paradigm for machine translation using recurrent  neural networks (RNNs) \cite{ChoMGBSB14}, \cite{GoogleNIC}, and \cite{KarpathyF14} use a deep convolutional neural network (CNN) to encode the input image, and a Long Short Term Memory (LSTM) \cite{HochreiterLSTM} RNN decoder to generate the output caption. These systems are trained end-to-end using back-propagation, and have achieved state-of-the-art results on MSCOCO. More recently in \cite{CapAttention}, the  use of spatial attention mechanisms on CNN layers to incorporate visual context---which implicitly conditions on the text generated so far---was incorporated into the  generation process. It has been shown and we have qualitatively observed that captioning systems that utilize attention mechanisms lead to better generalization, as these models can compose novel text descriptions based on the recognition of the global and local entities that comprise images.

As discussed in \cite{Ranzato}, deep generative models for text are typically trained to maximize the likelihood of the next ground-truth word given the previous \emph{ground-truth} word using back-propagation. This approach has been called ``Teacher-Forcing" \cite{ScheduledSampling}. However, this approach creates a mismatch between training and testing, since at test-time the model uses the previously generated words from the model distribution to predict the next word. This \emph{exposure bias} \cite{Ranzato}, results in error accumulation during generation at test time, since the model has never been exposed to its own predictions.

Several approaches to overcoming the exposure bias problem described above have recently been proposed. In \cite{ScheduledSampling} they show that feeding back the model's own predictions and slowly increasing the feedback probability $p$ during training leads to significantly better test-time performance.
Another line of work proposes ``Professor-Forcing'' \cite{ProfessorForcing}, a technique that uses adversarial training to encourage the dynamics of the recurrent network to be the same when training conditioned on ground truth previous words and when sampling freely from the network. 

While sequence models are usually trained using the cross entropy loss, they are typically evaluated at test time using discrete and non-differentiable NLP metrics such as BLEU \cite{BLEU}, ROUGE \cite{lin2004rouge}, METEOR \cite{banerjee2005meteor} or CIDEr \cite{CIDEr}. 
Ideally sequence models for image captioning should be trained to avoid  exposure bias \emph{and} directly optimize metrics for the task at hand.

Recently it has been shown that both the exposure bias and non-differentiable task metric issues can be addressed by incorporating techniques from Reinforcement Learning (RL) \cite{Sutton:1998}. Specifically in \cite{Ranzato}, Ranzato et al. use the REINFORCE algorithm \cite{Reinforce} to directly optimize non-differentiable, sequence-based test metrics, and overcome both issues. 
REINFORCE as we will describe, allows one to optimize the gradient of the expected reward by sampling from the model during training, and treating those samples as ground-truth labels (that are re-weighted by the reward they deliver). The major limitation of the approach is that the expected gradient computed using mini-batches under REINFORCE typically exhibit high variance, and without proper context-dependent normalization, is typically unstable. The recent discovery that REINFORCE  with proper bias correction using learned ``baselines" is effective has led to a flurry of work in applying REINFORCE to problems in RL, supervised learning, and variational inference \cite{schulman2015high, ZarembaS15, mnih2014neural}. Actor-critic methods \cite{Sutton:1998} , which instead train a second ``critic" network to provide an \emph{estimate} of the value of each generated word given the policy of an actor network, have also been investigated for sequence problems recently  \cite{actorCriticSequence}. These techniques overcome the need to sample from the policy's (actor's) action space, which can be enormous, at the expense of estimating future rewards, and training multiple networks based on one another's outputs, which as \cite{actorCriticSequence} explore, can also be unstable.

In this paper we present a new approach to sequence training which we call self-critical sequence training (SCST), and demonstrate that SCST can improve the performance of image captioning systems dramatically. SCST is a REINFORCE algorithm that, rather than estimating the reward signal, or how the reward signal should be normalized, utilizes the output of its own test-time inference algorithm to normalize the rewards it experiences. As a result, only samples from the model that outperform the current test-time system are given positive weight, and inferior samples are suppressed. Using SCST, attempting to estimate the reward signal, as actor-critic methods must do, and estimating normalization, as REINFORCE algorithms must do, is avoided, while at the same time harmonizing the model with respect to its test-time inference procedure. Empirically we find that directly optimizing the CIDEr metric with SCST and greedy decoding at test-time is highly effective. Our results on the MSCOCO evaluation sever establish a new state-of-the-art on the task, improving the best result in terms of CIDEr from 104.9 to 114.7.

\section{Captioning Models} \label{sec:captioner}

In this section we describe the recurrent models that we use for caption generation.\\

\noindent \textbf{FC models.} Similarly to \cite{GoogleNIC,KarpathyF14}, we first encode the input image $F$ using a deep CNN, and then embed it through a linear projection $W_{I}$ . Words are represented with one hot vectors that are embedded with a linear embedding $E$ that has the same output dimension as $W_{I}$.
The beginning of each sentence is marked with a special BOS token, and the end with an EOS token.
Under the model, words are generated and then fed back into the LSTM, with the image treated as the first word $W_{I} CNN(F)$. The following updates for the hidden units and cells of an LSTM define the model \cite{HochreiterLSTM}:\vspace{-0.1in} 
 \begin{align*}
x_{t}&=E 1_{w_{t-1}} \text{ for } t >1, x_1=W_{I} CNN(F)\\ 
i_{t}&= \sigma\left(W_{ix} x_{t}+W_{ih}h_{t-1}+ b_{i}\right) \;\;\, (\text {Input Gate})\\
f_{t}&=\sigma\left(W_{fx} x_{t}+W_{fh}h_{t-1}+ b_{f}\right) \; (\text {Forget Gate})\\\
o_{t}&=\sigma\left(W_{ox} x_{t}+W_{oh}h_{t-1}+ b_{o}\right) \;\, (\text {Output Gate})\\
c_{t}&= i_{t}\odot \phi(W^{\otimes}_{zx} x_{t}+W_{zh}^{\otimes}h_{t-1}+ b_{z}^{\otimes})+ f_{t}\odot c_{t-1} \\
h_t &= o_t \odot \tanh(c_{t}) \\
s_{t}&= W_{s} h_{t},
\vspace{-0.1in} 
\end{align*}
where $\phi$ is a maxout non-linearity with $2$ units ($\otimes$ denotes the units) and $\sigma$ is the sigmoid function. We initialize $h_0$ and $c_0$ to zero.  
The LSTM outputs a distribution over the next word $w_{t}$ using the softmax function:
\begin{equation}
w_{t}\sim \rm{softmax}\left(s_{t} \right)
\label{eq:dist}
\end{equation}
In our architecture, the hidden states and word and image embeddings have dimension $512$. 
Let $\theta$ denote the parameters of the model. Traditionally the parameters $\theta$ are learned by maximizing the likelihood of the observed sequence. 
Specifically, given a target ground truth sequence $(w^*_{1},\dots, w^*_{T})$, the objective is to minimize the cross entropy loss (XE):
\vspace{-0.12in}
\begin{equation}
L(\theta)=-\sum_{t=1}^{T} \log (p_{\theta}(w^*_{t}|w^*_{1},\dots w^*_{t-1})),
\label{eq:xent}
\end{equation}
where $p_{\theta}(w_t|w_1,\dots w_{t-1})$ is given by the parametric model in Equation \eqref{eq:dist}.\\

\noindent \textbf{Attention Model (Att2in).} Rather than utilizing a static, spatially pooled representation of the image, attention models dynamically re-weight the input spatial (CNN) features to focus on specific regions of the image at each time step. In this paper we consider a modification of the architecture of the attention model for captioning given in \cite{CapAttention}, and input the attention-derived image feature only to the cell node of the LSTM. 

\vspace{-0.1in}
\begin{align*}
x_{t}&=E 1_{w_{t-1}} \text{ for } t \geq 1~ w_0=BOS \\
i_{t}&= \sigma\left(W_{ix} x_{t}+W_{ih}h_{t-1}+ b_{i}\right)\;\;\;(\text {Input Gate})\\
f_{t}&=\sigma\left(W_{fx} x_{t}+W_{fh}h_{t-1}+ b_{f}\right)\; (\text {Forget Gate})\\\
o_{t}&=\sigma\left(W_{ox} x_{t}+W_{oh}h_{t-1}+ b_{o}\right)\;\,(\text {Output Gate})\\
c_{t}&= i_{t}\odot \phi(W_{zx}^{\otimes} x_{t}+\textcolor{red}{W_{zI}^{\otimes} I_{t}}+W_{zh}^{\otimes}h_{t-1}+ b_{z}^{\otimes})+ f_{t}\odot c_{t-1} \\
h_t &= o_t \odot \tanh(c_{t}) \\
s_{t}&= W_{s} h_{t},
\vspace{-0.2in} 
\end{align*}
where $I_{t}$ is the attention-derived image feature. This feature is derived as in \cite{CapAttention} as follows: given CNN features at $N$ locations $\{I_1,\dots I_{N}\}$, $I_{t}=\sum_{i=1}^N \alpha^i_t I_{i}$, where $\alpha_{t}={\rm softmax}(a_{t} + b_{\alpha})$, and $a^{i}_{t}=W\tanh(W_{aI}I_{i}+W_{ah}h_{t-1}+b_{a})$. In this work we set the dimension of $W$ to $1\times512$, and set $c_0$ and $h_0$ to zero. Let $\theta$ denote the parameters of the model. Then $p_{\theta}(w_t|w_1,\dots w_{t-1})$ is again defined by \eqref{eq:dist}. The parameters $\theta$ of attention models are also traditionally learned by optimizing the XE loss  \eqref{eq:xent}. 

\noindent \textbf{Attention Model (Att2all).} The standard attention model presented in \cite{CapAttention} also feeds then attention signal $I_{t}$ as an input into all gates of the LSTM, and the output posterior. In our experiments feeding  $I_{t}$ to all gates in addition to the input did not boost performance, but feeding  $I_{t}$ to both the gates and the outputs resulted in significant gains when ADAM \cite{ADAM} was used.

\section{Reinforcement Learning}
\noindent \textbf{Sequence Generation as an RL problem.} As described in the previous section, captioning systems are traditionally trained using the cross entropy loss. To directly optimize NLP metrics and address the exposure bias issue, we can cast our generative models in the Reinforcement Learning terminology as in \cite{Ranzato}. Our recurrent models (LSTMs) introduced above can be viewed as an ``agent'' that interacts with an external ``environment'' (words and image features). The parameters of the network, $\theta$, define a policy $p_{\theta}$, that results in an ``action'' that is the prediction of the next word.  After each action, the agent (the LSTM) updates its internal ``state'' (cells and hidden states of the LSTM, attention weights etc). Upon generating the end-of-sequence (EOS) token, the agent observes a ``reward'' that is, for instance, the CIDEr score of the generated sentence---we denote this reward by $r$. The reward is computed by an evaluation metric by comparing the generated sequence to corresponding ground-truth sequences. The goal of training is to minimize the negative expected reward:
\begin{equation}
L(\theta)= -  \mathbb{E}_{w^s\sim p_{\theta}} \left[ r(w^s)\right],
\end{equation}     
where $w^s= (w^s_1,\dots w^s_T)$ and  $w^s_{t}$ is the word sampled from the model at the time step $t$.
In practice $L(\theta)$ is typically estimated with a single sample from $p_{\theta}$:
\begin{equation*}
L(\theta) \approx-r(w^s),~ w^s \sim p_{\theta}.
\end{equation*}

\noindent \textbf{Policy Gradient with REINFORCE.} In order to compute the gradient $\nabla_{\theta}L(\theta)$, we use the REINFORCE algorithm \cite{Reinforce}(See also Chapter 13 in \cite{Sutton:1998}). REINFORCE is based on the observation that the expected gradient of a non-differentiable reward function can be computed as follows:
\begin{equation}
\nabla_{\theta}L(\theta)= - \mathbb{E}_{w^s \sim p_{\theta}} \left[ r(w^s) \nabla_{\theta} \log p_{\theta}(w^s) \right].
\end{equation}
In practice the expected gradient can be approximated using a single Monte-Carlo sample $w^s=(w^s_1 \dots w^s_{T})$ from $p_{\theta}$, for each training example in the minibatch:
\begin{equation*}
\nabla_{\theta}L(\theta)\approx- r(w^s) \nabla_{\theta} \log p_{\theta}(w^s).
\end{equation*}

\noindent \textbf{REINFORCE with a Baseline.} The policy gradient given by REINFORCE can be generalized to compute the reward associated with an action value \emph{relative} to a reference reward or \emph{baseline} $b$:
\begin{equation}
\nabla_{\theta}L(\theta)= - \mathbb{E}_{w^s \sim p_{\theta}} \left[ (r(w^s)- b ) \nabla_{\theta} \log p_{\theta}(w^s)\right].
\end{equation}
The baseline can be an arbitrary function, as long as it does not depend on the ``action''  $w^s$ \cite{Sutton:1998}, since in this case:
\begin{align}
\mathbb{E}_{w^s\sim p_{\theta}} \left[ b  \nabla_{\theta} \log p_{\theta}(w^s)\right]& = b \sum_{w_s} \nabla_{\theta} p_{\theta}(w^s)  \nonumber\\
&=b \nabla_{\theta}  \sum_{w_s}  p_{\theta}(w^s)  \nonumber\\
&=  b  \nabla_{\theta} 1 =0.
\label{eq:zerobiasGradient}
\end{align}
This shows that the baseline does not change the expected gradient, but importantly, it can reduce the variance of the gradient estimate. 
For each training case, we again approximate the expected gradient with a single sample $w^s\sim p_{\theta}$:
\begin{equation}
\nabla_{\theta}L(\theta)\approx- (r(w^s)-b) \nabla_{\theta} \log p_{\theta}(w^s).
\label{eq:zerobiasGradientSample}
\end{equation}
Note that if $b$ is function of $\theta$ or $t$ as in \cite{Ranzato}, equation \eqref{eq:zerobiasGradient} still holds and $b(\theta)$ is a valid baseline. \\

\noindent \textbf{Final Gradient Expression.} Using the chain rule, and the parametric model of $p_{\theta}$ given in Section \ref{sec:captioner} we have: 
$$\nabla_{\theta}L(\theta)=\sum_{t=1}^{T} \frac{\partial L(\theta)}{\partial s_{t}}\frac{\partial s_{t}}{\partial \theta},$$
where $s_{t}$ is the input to the softmax function. Using REINFORCE with a baseline $b$ the estimate of  the gradient of $\frac{\partial L(\theta)}{\partial s_{t}}$ is given by \cite{ZarembaS15}:
\begin{equation}
\frac{\partial L(\theta)}{\partial s_{t}}\approx (r(w^s)-b)(p_{\theta}(w_{t}|h_t)- 1_{w^s_{t}}).
\label{eq:GradientREINFORCE}
\end{equation}

\section{Self-critical sequence training (SCST)}

The central idea of the self-critical sequence training (SCST) approach is to baseline the REINFORCE algorithm with the reward obtained by the current model under the inference algorithm used at test time. The gradient of the negative reward of a sample $w^s$ from the model w.r.t. to the softmax activations at time-step $t$ then becomes: 
\begin{equation}
\frac{\partial L(\theta)}{\partial s_{t}}=(r(w^s)-r(\hat{w}))(p_{\theta}(w_{t}|h_t)- 1_{w^s_{t}}).
\end{equation}
where $r(\hat{w})$ again is the reward obtained by the current model under the inference algorithm used at test time.
Accordingly, samples from the model that return higher reward than $\hat{w}$ will be ``pushed up", or increased in probability, while samples which result in lower reward will be suppressed. Like MIXER \cite{Ranzato}, SCST has all the advantages of REINFORCE algorithms, as it directly optimizes the true, sequence-level, evaluation metric, but avoids the usual scenario of having to learn a (context-dependent) \emph{estimate} of expected future rewards as a baseline. In practice we have found that SCST has much lower variance, and can be more effectively trained on mini-batches of samples using SGD. 
\begin{figure*}[ht!]
\begin{center}
\includegraphics[width=0.8\linewidth]{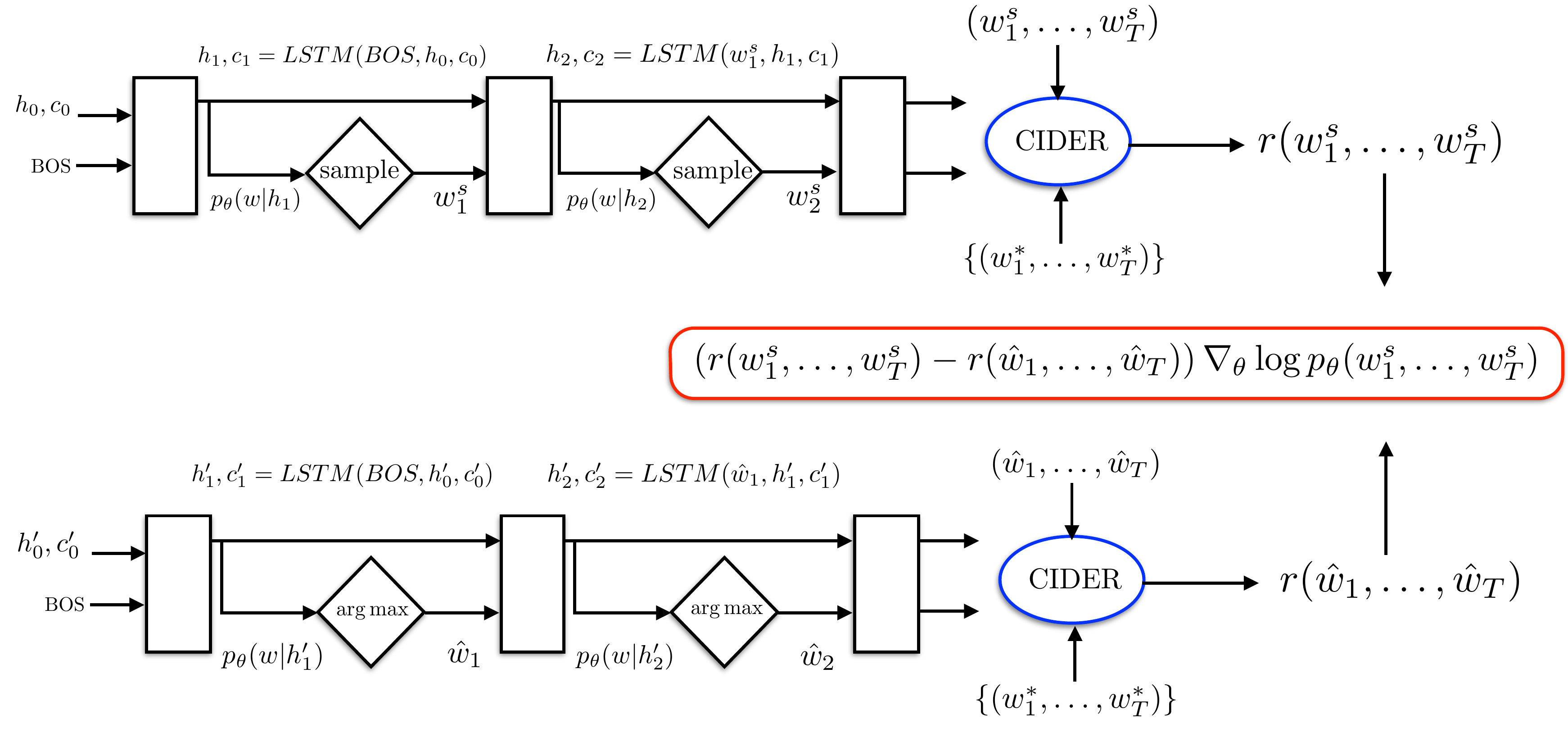}
\end{center}
   \caption{Self-critical sequence training (SCST). The weight put on words of a sampled sentence from the model is determined by the difference between the reward for the sampled sentence and the reward obtained by the estimated sentence under the test-time inference procedure (greedy inference depicted). This harmonizes learning with the inference procedure, and lowers the variance of the gradients, improving the training procedure.}
\label{fig:short}
\end{figure*}
Since the SCST baseline is based on the test-time estimate under the current model, SCST is forced to improve the performance of the model under the inference algorithm used at test time. This encourages training/test time consistency like the maximum likelihood-based approaches  ``Data as Demonstrator" \cite{ScheduledSampling},  ``Professor Forcing" \cite{ProfessorForcing}, and E2E \cite{Ranzato}, but importantly, it can directly optimize sequence metrics. Finally, SCST is self-critical, and so avoids all the inherent training difficulties associated with actor-critic methods, where a second ``critic" network must be trained to estimate value functions, and the actor must be trained on \emph{estimated} value functions rather than actual rewards.

In this paper we focus on scenario of greedy decoding, where:
\begin{equation}
\hat{w}_t = \arg \max_{w_t} p(w_t\,|\,h_t)
\end{equation}
This choice, depicted in Figure \ref{fig:short}, minimizes the impact of baselining with the test-time inference algorithm on training time, since it requires only one additional forward pass, and trains the system to be optimized for fast, greedy decoding at test-time.
%
%

\noindent \textbf{Generalizations.}
The basic SCST approach described above can be generalized in several ways. 

One generalization is to condition the baseline on what has been generated (i.e. sampled) so far, which makes the baseline \emph{word-dependent}, and further reduces the variance of the reward signal by making it dependent only on \emph{future rewards}. This is achieved by baselining the reward for word $w^s_t$ at timestep $t$ with the reward obtained by the word sequence $\bar{w} = \{w^s_{1:t-1}, \hat{w}_{t:T}\}$, which is generated by sampling tokens for timesteps $1:t-1$, and then executing the inference algorithm to complete the sequence. The resulting reward signal, $r(w^s)-r(\bar{w})$, is a baselined future reward (advantage) signal that conditions on both the input image and the sequence $w^s_{1:t-1}$, and remains unbiased. We call this variant \emph{time-dependent SCST (TD-SCST)}.

Another important generalization is to utilize the inference algorithm as a critic to replace the learned critic of traditional actor-critic approaches. Like for traditional actor-critic methods, this biases the learning procedure, but can be used to trade off variance for bias. Specifically, the \emph{primary} reward signal at time $t$ can be based on a sequence that samples only $n$ future tokens, and then executes the inference algorithm to complete the sequence. The primary reward is then based on $\tilde{w} = \{w^s_{1:t+n}, \hat{w}_{t+n+1:T}\}$, and can further be baselined in a time-dependent manner using TD-SCST. The resulting reward signal in this case is $r(\tilde{w} )-r(\bar{w})$. We call this variant \emph{True SCST}.

We have experimented with both TD-SCST and ``True" SCST as described above on the MSCOCO task, but found that they did not lead to significant additional gain. We have also experimented with learning a control-variate for the SCST baseline on MSCOCO to no avail. Nevertheless, we anticipate that these generalizations will be important for other sequence modeling tasks, and policy-gradient-based RL more generally.

\section{Experiments}

\noindent \textbf{Dataset.} We evaluate our proposed method on the  MSCOCO dataset \cite{MSCOCO}. For offline evaluation purposes we used the data splits from  \cite{Karpathy}. The training set contains $113,287$ images, along with $5$ captions each. We use a set of $5K$ image for validation and report results  on a test set of 
$5K$ images as well,  as given in \cite{Karpathy}. We report four  widely used automatic evaluation metrics, BLEU-4, ROUGEL, METEOR, and CIDEr. We prune the vocabulary and drop any word that has count less then five, we end up with a vocabulary of size 10096 words.\\

\noindent \textbf{Image Features}
1) \emph{FC Models.} We use two type of Features: a) (FC-2k) features, where we encode each image with Resnet-101 ($101$ layers) \cite{he15deepresidual}. Note that we do not rescale or crop each image. Instead we encode the full image with the final convolutional layer of resnet, and apply average pooling, which results in a vector of dimension $2048$. b) (FC-15K) features where we  stack average pooled  $13$ layers  of Resnet-101 ($11\times1024$ and $2\times2048$). These $13$ layers are the odd layers of conv4 and conv5, with the exception of the 23rd layer of conv4, which was omitted. This results in a feature vector of dimension 15360.\\
2) \emph{Spatial  CNN features for Attention models:} (Att2in) We encode each image using the residual convolutional neural network Resnet-101 \cite{he15deepresidual}. Note that we do not rescale or crop the image. Instead we encode the full image with the final convolutional layer of Resnet-101, and apply spatially adaptive max-pooling so that the output has a fixed  size of $14\times 14 \times 2048$. At each time step the attention model produces an attention mask over the $196$ spatial locations. This mask is applied and then the result is spatially averaged to produce a $2048$ dimension representation of the attended portion of the image. \\

\noindent \textbf{Implementation Details.} The LSTM hidden, image, word and attention embeddings dimension are fixed to 512 for all of the models discussed herein. All of our models are trained according to the following recipe, except where otherwise noted. We initialize all models by training the model under the XE objective using ADAM \cite{ADAM} optimizer with an initial learning rate of $5\times 10^{-4}$. We anneal the learning rate by a factor of $0.8$ every three epochs, and increase the probability of feeding back a sample of the word posterior by $0.05$ every 5 epochs until we reach a feedback probability $0.25$ \cite{ScheduledSampling}. We evaluate at each epoch the model on the development set and select the model with best CIDEr score as an initialization for SCST training. We then run SCST training initialized with the XE model to optimize the CIDEr metric (specifically, the CIDEr-D metric) using ADAM with a learning rate $5\times10^{-5}$ \footnote{In the case of the Att2all models, the XE model was trained for only 20 epochs, and the learning rate was also annealed during RL training.}. 
Initially when experimenting with FC-2k and FC-15k models we utilized curriculum learning (CL) during training, as proposed in \cite{Ranzato}, by increasing the number of words that are sampled and trained under CIDEr by one each epoch (the prefix of the sentence remains under the XE criterion until eventually being subsumed). We have since realized that for the MSCOCO task CL is not required, and provides little to no boost in performance. The results reported here for the FC-2K and FC-15K models are trained with CL, while the attention models were trained directly on the entire sentence for all epochs after being initialized by the XE seed models.

\begin{table}[ht]
\begin{center}
\begin{tabular}{l|cccc}
\hline \hline
Training & \multicolumn{4}{c}{Evaluation Metric} \\
Metric & CIDEr & BLEU4 & ROUGEL & METEOR  \\
\hline
XE & 90.9 & 28.6 & 52.3 & 24.1  \\ 
XE (beam) & 94.0 & 29.6 & 52.6 & 25.2 \\ 
\hline

MIXER-B & 101.9 & 30.9 & 53.8 & 24.9 \\ 
MIXER & 104.9 & 31.7 & {\bf 54.3} &  25.4 \\
SCST  & {\bf 106.3} & {\bf 31.9} & {\bf 54.3} & {\bf 25.5}  \\ 
\hline \hline
\end{tabular}
\end{center}
\caption{Performance of self-critical sequence training (SCST) versus MIXER \cite{Ranzato} and MIXER without a baseline (MIXER-B) on the test portion of the Karpathy splits when trained to optimize the CIDEr metric (FC-2K models). All improve the seed cross-entropy trained model, but SCST outperforms MIXER.}
\label{rlcompare_table}
\end{table}
\vspace{-0.1in}
\begin{table}[th]
\begin{center}
\small\addtolength{\tabcolsep}{-2pt} 
\begin{tabular}{l|lrl|lrl|lrl}
\hline\hline
Metric &  \multicolumn{3}{c|}{REINFORCE} &   \multicolumn{3}{c|}{MIXER-CL} & \multicolumn{3}{c}{SCST} \\
\hline 
CIDEr				& 110.4&$\pm$& 0.5 	& 113.2&$\pm$& 0.2 		&	 \bf 113.8&$\pm$& 0.3 	\\  
BLEU4	   			& 32.8 &$\pm$& 0.1 		& 33.8&$\pm$&0.2		&	\bf	34.1 &$\pm$& 0.1 	\\
ROUGE			& 55.2 &$\pm$& 0.1		& 55.6&$\pm$&0.1		&	\bf	55.7 &$\pm$& 0.1 	\\
METEOR			& 26.0 &$\pm$& 0.04	& 26.5&$\pm$&0.1		&	 \bf 26.6 &$\pm$& 0.04  	\\
\hline\hline
\end{tabular}
\caption{Mean/std. performance of SCST versus REINFORCE and REINFORCE with learned baseline (MIXER less CL), for Att2all models (4 seeds) on the Karpathy test set (CIDEr optimized). A one-sample t-test on the gain of SCST over MIXER less CL rejects the null hypothesis on all metrics except ROUGE for $\alpha=0.1$ (i.e. $p_{null} < 0.1$).}
\label{rlcompare_table2}
\end{center}
\end{table}
\vspace{-0.15in}

\subsection{Offline Evaluation}

\noindent \textbf{Evaluating different RL training strategies.}

\begin{figure}[ht]
\begin{center}
\includegraphics[angle=0, origin=c, width=0.5\textwidth]{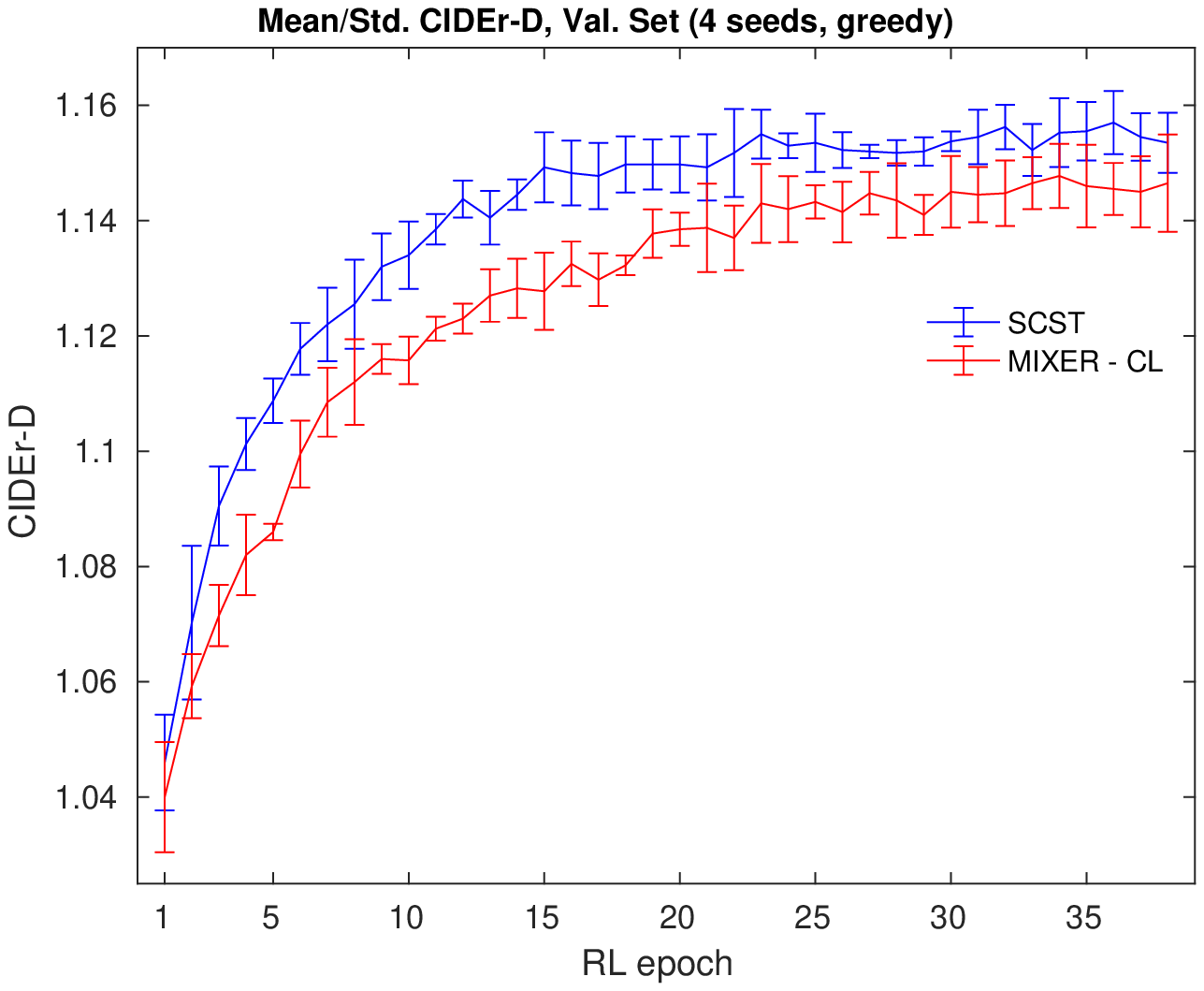}
\end{center}
\caption{Mean/std. CIDEr of SCST versus REINFORCE with learned baseline (MIXER less CL) with greedy decoding, for Att2all models (4 seeds) on the Karpathy validation set (CIDEr-D optimized).}
\label{fig:scst_vs_mixer_greedy}
\end{figure}

\begin{figure}[ht]
\begin{center}
\includegraphics[angle=0, origin=c, width=0.5\textwidth]{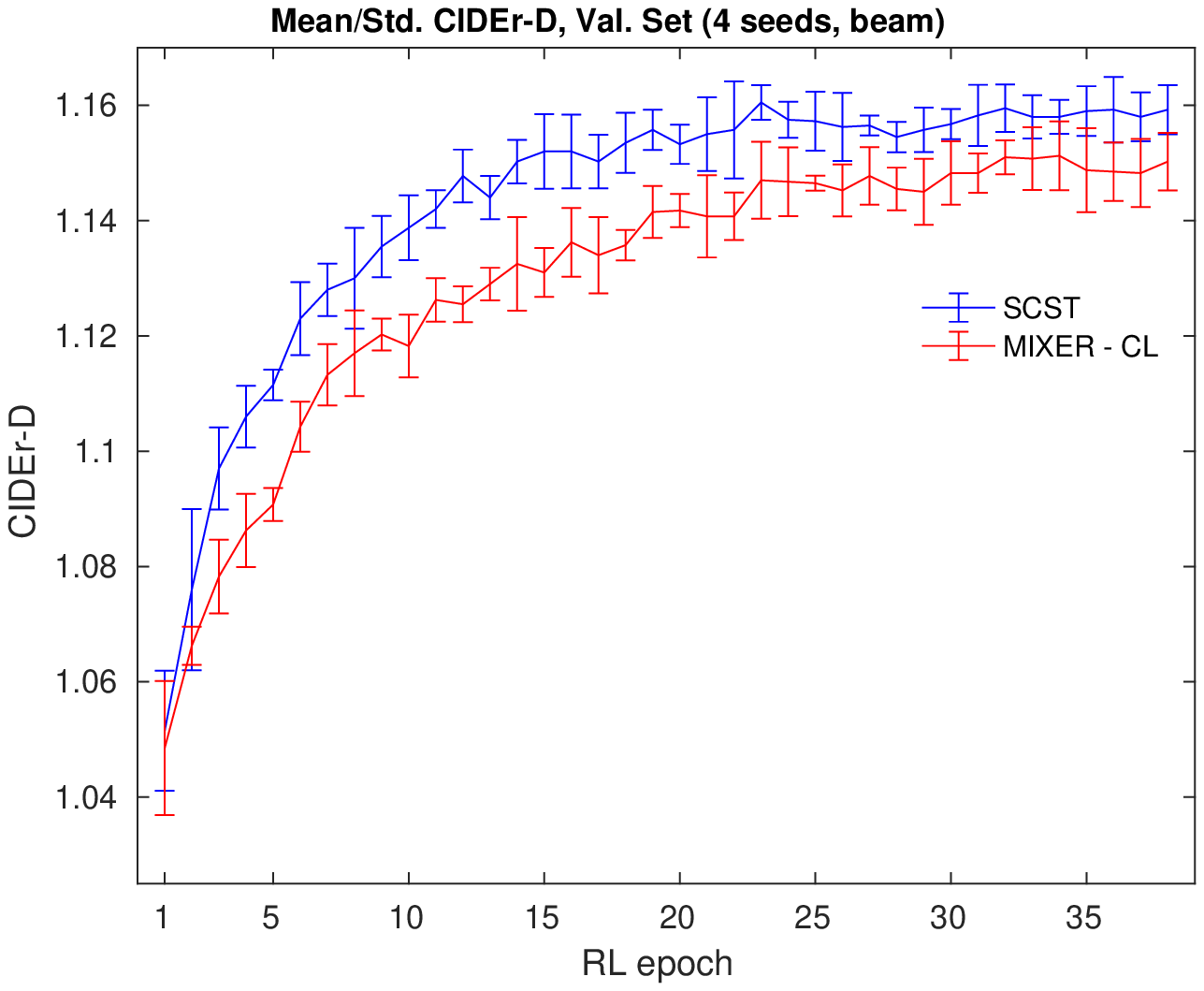}
\end{center}
\caption{Mean/std. CIDEr of SCST versus REINFORCE with learned baseline (MIXER less CL) with beam search decoding, for Att2all models (4 seeds) on the Karpathy validation set (CIDEr-D optimized).}
\label{fig:scst_vs_mixer_beam}
\end{figure}

\begin{figure}[ht]
\begin{center}
\includegraphics[angle=0, origin=c, width=0.5\textwidth]{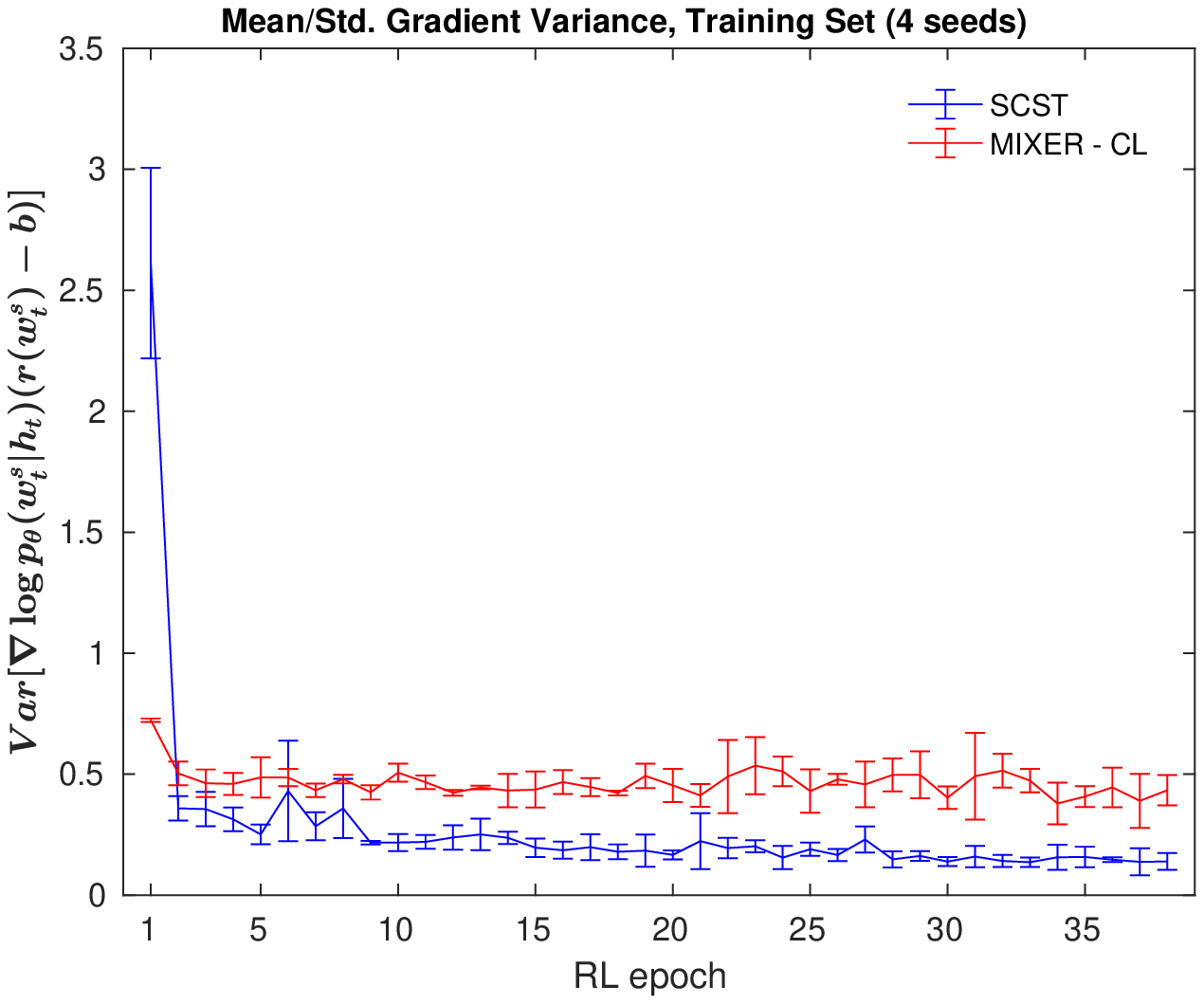}
\end{center}
\caption{Mean/std. gradient variance of SCST versus REINFORCE with learned baseline (MIXER less CL), for Att2all models (4 seeds) on the Karpathy training set (CIDEr-D optimized).}
\label{fig:scst_vs_mixer_gradvar}
\end{figure}

\begin{figure}[ht]
\begin{center}
\includegraphics[angle=0, origin=c, width=0.5\textwidth]{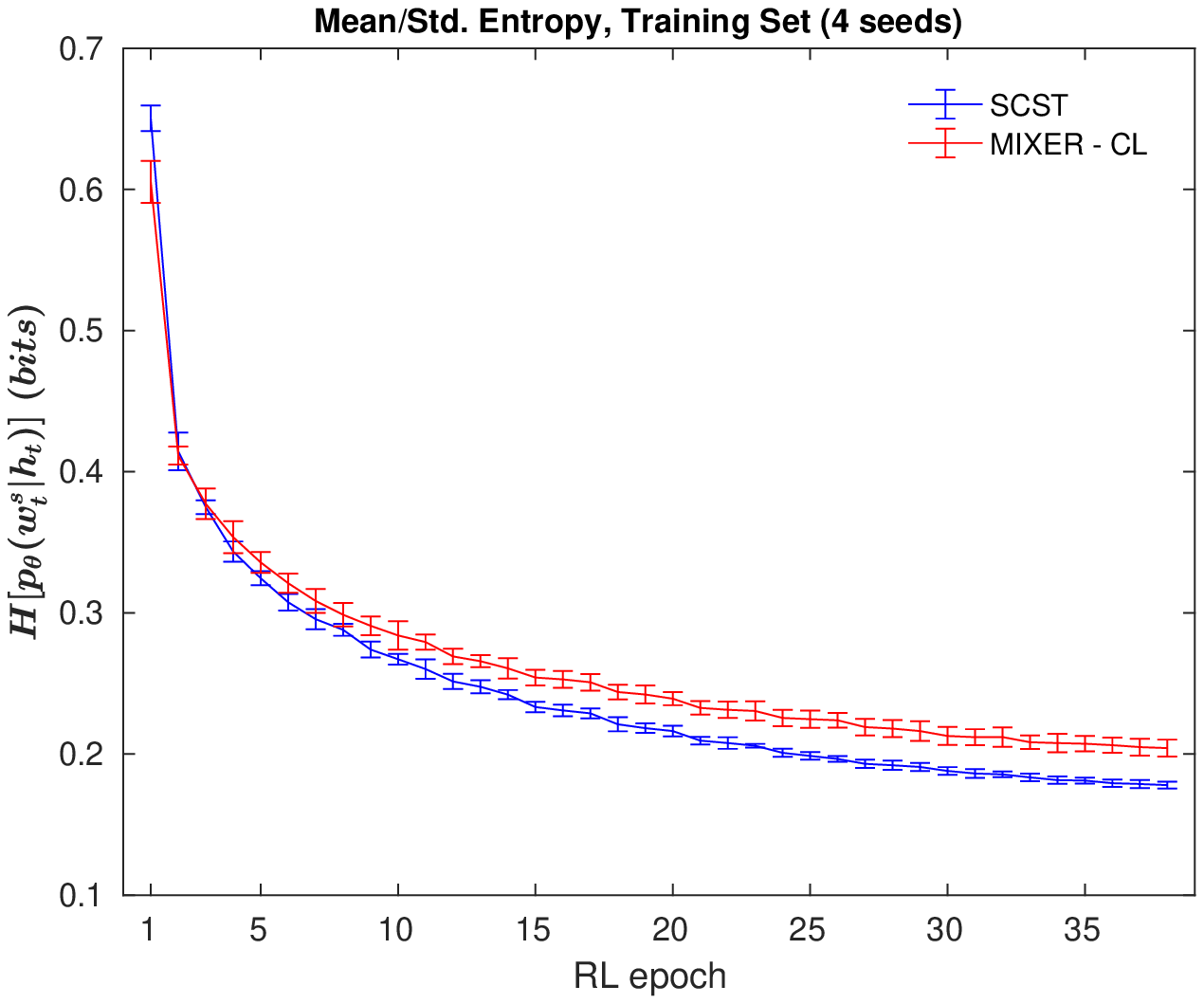}
\end{center}
\caption{Mean/std. word posterior entropy of SCST versus REINFORCE with learned baseline (MIXER less CL), for Att2all models (4 seeds) on the Karpathy training set (CIDEr-D optimized).}
\label{fig:scst_vs_mixer_post_entropy}
\end{figure}

Table \ref{rlcompare_table} compares the performance of SCST to MIXER \cite{Ranzato} (test set, Karpathy splits). In this experiment, we utilize ``curriculum learning" (CL) by optimizing the expected reward of the metric on the last $n$ words of each training sentence, optimizing XE on the remaining sentence prefix, and slowly increasing $n$. The results reported were generated with the optimized CL schedule reported in  \cite{Ranzato}. We found that CL was not necessary to train both SCST and REINFORCE with a learned baseline on MSCOCO---turning off CL sped up training and yielded equal or better results. The gain of SCST over learned baselines was consistent, regardless of the CL schedule and the initial seed. Figures \ref{fig:scst_vs_mixer_greedy} and \ref{fig:scst_vs_mixer_beam} and table \ref{rlcompare_table2} compare the performance of SCST and MIXER less CL for Att2all models on the Karpathy validation and test splits, respectively. Figure \ref{fig:scst_vs_mixer_gradvar} and \ref{fig:scst_vs_mixer_post_entropy} further compare their gradient variance and word posterior entropy on the Karpathy training set. While both techniques are unbiased, SCST in general has lower gradient variance, which translates to improved training performance. Interestingly, SCST has much \emph{higher} gradient variance than MIXER less CL during the first epoch of training, as most sampled sentences initially score significantly lower than the sentence produced by the test-time inference algorithm.

\vskip 0.1in
\noindent \textbf{Training on different metrics.}

We experimented with training directly on the evaluation metrics of the MSCOCO challenge. Results for FC-2K models are depicted in table \ref{metric_table}. 
In general we can see that optimizing for a given metric during training leads to the best performance on that same metric at test time, an expected result. 
We experimented with training on multiple test metrics, and found that we were unable to outperform the overall performance of the model trained only on the CIDEr metric, which lifts the performance of all other metrics considerably. For this reason most of our experimentation has since focused on optimizing CIDEr.\\



\begin{table}[ht]
\begin{center}
\begin{tabular}{l|cccc}
\hline \hline
Training & \multicolumn{4}{c}{Evaluation Metric} \\
Metric & CIDEr & BLEU4 & ROUGEL & METEOR  \\
\hline
XE & 90.9 & 28.6 & 52.3 & 24.1  \\ 
XE (beam) & 94.0 & 29.6 & 52.6 & 25.2 \\ 
\hline
CIDEr      & {\bf 106.3} & 31.9 & 54.3 & 25.5  \\ 
BLEU       & 94.4 & {\bf 33.2} & 53.9 & 24.6  \\ 
ROUGEL & 97.7 & 31.6 & {\bf 55.4} & 24.5  \\ 
METEOR & 80.5 & 25.3 & 51.3 & {\bf 25.9}  \\ 

\hline \hline
\end{tabular}
\end{center}
\caption{Performance on the test portion of the Karpathy splits \cite{Karpathy} as a function of training metric ( FC-2K models). Optimizing the CIDEr metric increases the overall performance under the evaluation metrics the most significantly. The performance of the seed cross-entropy (XE) model is also depicted. All models were decoded greedily, with the exception of the XE beam search result, which was optimized to beam 3 on the validation set.}
\label{metric_table}
\end{table}

\begin{table}[th]
\small\addtolength{\tabcolsep}{-2pt} 
\begin{tabular}{l|c|cccc}
\hline \hline
\multicolumn{6}{c}{\multirow{2}{*}{Single Best Models (XE)}} \\
\multicolumn{6}{c}{}\\
\hline
Model &  Search & \multicolumn{4}{c}{Evaluation Metric} \\
  Type & Method & CIDEr & BLEU4 & ROUGEL & METEOR  \\
\hline
\multirow{2}{*}{FC-2K}   & greedy & 90.9 & 28.6 & 52.3 & 24.1  \\ 
& beam & 94.0 & 29.6 & 52.6 & 25.2 \\ 
\hline
\multirow{2}{*}{FC-15K} & greedy & 94.1  & 29.5  & 52.9  & 24.4  \\ 
				   & beam & 96.1  & 30.0  & 52.9  & 25.2  \\ 
\hline
\multirow{2}{*}{Att2in} & greedy & 99.0 & 30.6 & 53.8 & 25.2  \\ 
& beam & {\bf 101.3} & {\bf 31.3} & {\bf 54.3} &{\bf  26.0}  \\ 
\hline
{Att2all} & greedy & 97.9 & 29.3 & 53.4 & 25.4 \\
(RL seed) & beam & 99.4 & 30.0 & 53.4 & 25.9 \\

\end{tabular}
\begin{tabular}{l|c|cccc}
\hline \hline
\multicolumn{6}{c}{\multirow{2}{*}{Single Best Models (SCST unless noted o.w.)}} \\
\multicolumn{6}{c}{}\\
\hline
Model &  Search & \multicolumn{4}{c}{Evaluation Metric} \\
Type  &Method& CIDEr & BLEU4 & ROUGEL & METEOR  \\
\hline
\multirow{2}{*}{FC-2K}   & greedy &  106.3 & 31.9 & 54.3 & 25.5  \\ 
& beam & 106.3 & 31.9 & 54.3 & 25.5  \\ 
\hline
\multirow{2}{*}{FC-15K} & greedy & 106.4 & 32.2 & 54.6 & 25.5  \\ 
& beam & 106.6 & 32.4 & 54.7 & 25.6  \\ 
\hline
\multirow{2}{*}{Att2in}   &  greedy & 111.3 & 33.3 & 55.2 &  26.3  \\ 
& beam & 111.4 & 33.3 &  55.3 &  26.3 \\ 
\hline
4 Att2all & greedy & 110.2 & 32.7 & 55.1 & 26.0 \\
(REINF.)& beam & 110.5 & 32.8 & 55.2 & 26.1 \\
\hline
4 Att2all & greedy & 112.9 & 34.0 & 55.5 & 26.4 \\ 
(MIXER-CL)& beam &113.0 & 34.1 & 55.5 & 26.5 \\
\hline
\multirow{2}{*}{Att2all} & greedy & 113.7 & 34.1 & 55.6 & 26.6 \\
& beam & {\bf 114.0} & {\bf 34.2} & {\bf 55.7} & {\bf 26.7} \\
\hline \hline
\end{tabular}

\caption{Performance of the best XE and corr. SCST-trained single models on the Karpathy test split (best of 4 random seeds). The results obtained via the greedy decoding and optimized beam search are depicted. Models learned using SCST were trained to directly optimize the CIDEr metric. MIXER less CL results (MIXER-) are also included.}
\label{tab:SingFcVsAtt}
\end{table}

\noindent \textbf{Single FC-Models Versus Attention Models.}  We trained  FC models (2K and 15 K), as well as  attention models  (Att2in and Att2all) using SCST with the CIDEr metric. We trained $4$ different models for each FC and attention type, starting the optimization  from four  different  random seeds \footnote{pls. consult the supp. material for further details on model training.}. We report in Table \ref{tab:SingFcVsAtt}, the system with best performance for each family of models on the test portion of Karpathy splits \cite{Karpathy}. We see that the FC-15K models outperform the FC-2K models. Both FC models are outperformed by the attention models, which  establish a new state of the art for a single model performance on Karpathy splits. Note that this quantitative evaluation favors attention models is inline with our observation that attention models tend to generalize better and compose outside of the context of the training of MSCOCO, as we will see in Section \ref{Sec:Example}.

\begin{table}[t]

\small\addtolength{\tabcolsep}{-2.0pt} 
\begin{tabular}{l|c|cccc}
\hline \hline
\multicolumn{6}{c}{\multirow{2}{*}{Ensembled Models (XE)}} \\
\multicolumn{6}{c}{}\\
\hline
Model &  Search & \multicolumn{4}{c}{Evaluation Metric} \\
 Type & Method & CIDEr & BLEU4 & ROUGEL & METEOR  \\

\hline
\multirow{2}{*}{4 FC-2K } & greedy & 96.3 & 30.1 & 53.5 & 24.8 \\
	& beam  & 99.2 &  31.2 & 53.9 & 25.8  \\ 
\hline
 \multirow{2}{*}{4 FC-15K } & greedy & 97.7 & 30.7 & 53.8 & 25.0 \\
 &  beam & 100.7 &  31.7 & 54.2 &  26.0 \\ 
\hline
\multirow{2}{*}{4 Att2in } & greedy & 102.8 & 32.0 & 54.7 & 25.7 \\ 
 & beam & \bf{106.5} & \bf{32.8} & \bf{55.1} & \bf{26.7}  \\ 
\hline
Att2all & greedy & 102.0 & 31.2 & 54.4 & 26.0 \\
 (RL seeds) & beam & 104.7 & 32.2 & 54.8 & 26.7 \\


\hline \hline
\multicolumn{6}{c}{\multirow{2}{*}{Ensembled Models (SCST unless o.w. noted)}} \\
\multicolumn{6}{c}{}\\
\hline
Model &  Search & \multicolumn{4}{c}{Evaluation Metric} \\
 Type & Method & CIDEr & BLEU4 & ROUGEL & METEOR  \\

\hline
\multirow{2}{*}{4 FC-2K } & greedy & 108.9 & 33.1 & 54.9 & 25.7 \\
& beam & 108.9 & 33.2 & 54.9 & 25.7 \\
\hline
\multirow{2}{*}{4 FC-15K  } & greedy & 110.4 & 33.4 & 55.4 & 26.1 \\
& beam & 110.4 & 33.4 & 55.4 & 26.2 \\ 
\hline
\multirow{2}{*}{4 Att2in } & greedy & 114.7 & 34.6 & 56.2 & 26.8 \\ 
& beam & 115.2 & 34.8 & 56.3 & 26.9  \\ 
\hline
4 Att2all & greedy & 113.8 & 34.2 & 56.0 & 26.5 \\
(REINF.)& beam &113.6 & 33.9 & 55.9 & 26.5 \\
\hline
4 Att2all   & greedy & 116.6 & 34.9 & 56.3 & 26.9 \\
(MIXER-CL) & beam & 116.7 & 35.1  & 56.4 & 26.9 \\
\hline
\multirow{2}{*}{4 Att2all } & greedy & 116.8 & 35.2 & 56.5 & 27.0 \\
& beam & \bf 117.5 & \bf 35.4 & \bf 56.6 & \bf 27.1 \\
\hline
\hline
\end{tabular}
\caption{Performance of Ensembled XE and SCST-trained models on the Karpathy test split (ensembled over 4 random seeds). The models learned using self-critical sequence training (SCST) were trained to optimize the CIDEr metric. MIXER less CL results (MIXER-) are also included.}
\label{tab:EnsembleFcVsAtt}
\end{table}

\noindent \textbf{Model Ensembling.} In this section we investigate the performance of ensembling over 4 random seeds of the XE and SCST-trained FC models and attention models. We see in Table \ref{tab:EnsembleFcVsAtt} that ensembling improves performance and confirms the supremacy of attention modeling, and establishes yet another state of the art result on Karpathy splits \cite{Karpathy}. Note that in our case we ensemble only $4$ models and we don't do any fine-tuning of the Resnet. NIC \cite{VinyalsTBE16}, in contrast, used an ensemble of $15$ models with fine-tuned CNNs. 

\vspace{-0.00in}
\subsection{Online Evaluation on MS-COCO Server}

Table \ref{tab:serverresults} reports the performance of two variants of 4 ensembled attention models trained with self-critical sequence training (SCST) on the official MSCOCO evaluation server. The previous best result on the leaderboard (as of April 10, 2017) is also depicted. We outperform the previous best system on all evaluation metrics.

\begin{table}[th]
\small\addtolength{\tabcolsep}{-0.0pt} 
\begin{tabular}{l|cccc}
\hline \hline
Ensemble  & \multicolumn{4}{c}{Evaluation Metric} \\
 SCST models & CIDEr & BLEU4 & ROUGEL & METEOR  \\
\hline
Ens. 4 (Att2all) & \bf{114.7} & \bf{35.2} & \bf{56.3} & \bf{27.0} \\
Ens. 4 (Att2in) & {112.3} & {34.4}  & {55.9} & {26.8}  \\ 
Previous best & 104.9 & 34.3 & 55.2 & 26.6 \\
\hline \hline
\end{tabular}
\caption{Performance of 4 ensembled attention models trained with self-critical sequence training (SCST) on the official MSCOCO evaluation server (5 reference captions). The previous best result on the leaderboard (as of 04/10/2017) is also depicted ( \href{http://mscoco.org/dataset/\#captions-leaderboard}{http://mscoco.org/dataset/\#captions-leaderboard}, Table C5, Watson Multimodal).}
\label{tab:serverresults}
\end{table}
\vspace{-0.1in}

\section{Example of Generated Captions}\label{Sec:Example}

\begin{figure*}[ht]
\vspace{-0.2in}
\begin{center}
 \includegraphics[width=0.4\linewidth]{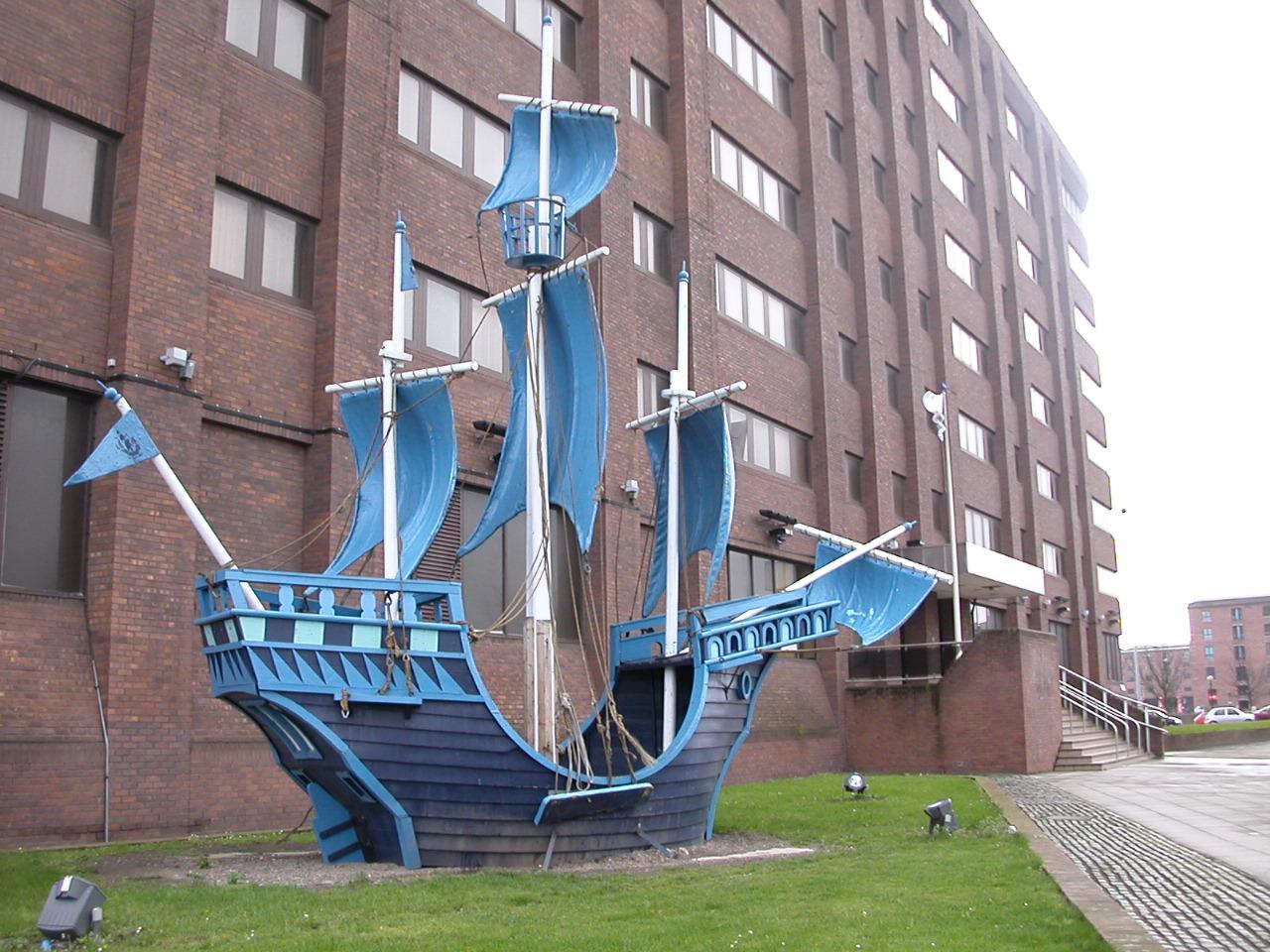}
\end{center}
   \caption{An image from the objects out-of-context (OOOC) dataset of images from \cite{Jinchoi_contextmodels}.}
\label{fig:boat}
\end{figure*}
\begin{figure*}[ht]
\begin{center}
 \includegraphics[width=0.9\linewidth]{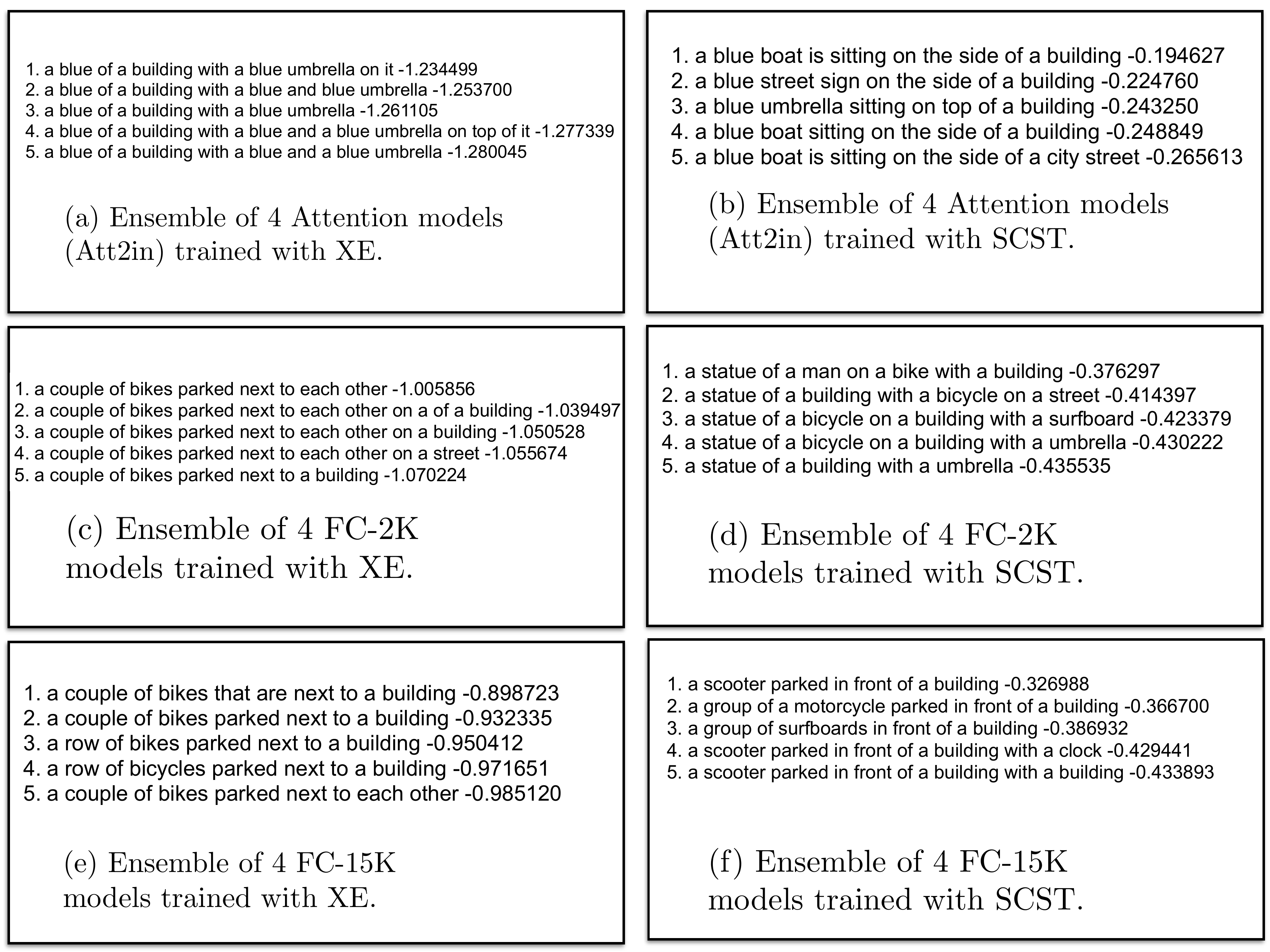}
\end{center}
   \caption{Captions generated for the image depicted in Figure \ref{fig:boat} by the various models discussed in the paper. Beside each caption we report the average log probability of the words in the caption. On this image, which presents an object situated in an atypical context \cite{Jinchoi_contextmodels}, the FC models fail to give an accurate description, while the attention models handle the previously unseen image composition well. The models trained with SCST return a more accurate and more detailed summary of the image.}
\label{fig:CapGen4}
\vspace{0.25in}
\end{figure*}

Here we provide a qualitative example of the captions generated by our systems for the image in figure \ref{fig:boat}. 
This picture is taken from the objects out-of-context (OOOC) dataset of images \cite{Jinchoi_contextmodels}. It depicts a boat situated in an unusual context, and tests the ability of our models to compose descriptions of images that differ from those seen during training. The top 5 captions returned by the XE and SCST-trained FC-2K, FC-15K, and attention model ensembles when deployed with a decoding ``beam" of 5 are depicted in figure \ref{fig:CapGen4} \footnote{pls. consult the the supp. material for further details on beam search.}.
On this image the FC models fail completely, and the SCST-trained ensemble of attention models is the only system that is able to correctly describe the image. In general we found that the performance of all captioning systems on MSCOCO data is qualitatively similar, while on images containing objects situated in an uncommon context \cite{Jinchoi_contextmodels} (i.e. unlike the MSCOCO training set) our attention models perform much better, and SCST-trained attention models output yet more accurate and descriptive captions. 
In general we qualitatively found that SCST-trained attention models describe images more accurately, and with higher confidence, as reflected in Figure  \ref{fig:CapGen4}, where the average of the log-likelihoods of the words in each generated caption are also depicted. Additional examples can be found in the supplementary material.  Note that we found that Att2in attention models actually performed better than our Att2all models when applied to images ``from the wild", so here we focus on demonstrating them.

\section{Discussion and Future Work}

In this paper we have presented a simple and efficient approach to more effectively baselining the REINFORCE algorithm for policy-gradient based RL, which allows us to more effectively train on non-differentiable metrics, and leads to significant improvements in captioning performance on MSCOCO---our results on the MSCOCO evaluation sever establish a new state-of-the-art on the task. The self-critical approach, which normalizes the reward obtained by sampled sentences with the reward obtained by the model under the test-time inference algorithm is intuitive, and avoids having to estimate both action-dependent and action-independent reward functions. 

{\small
\bibliographystyle{unsrt}
\bibliography{cap_rl_arxiv_Extended_tables}
}


\onecolumn 
\appendix
\begin{center}
\bf \Large Self-critical Sequence Training for Image Captioning:\\
 Supplementary Material
\end{center}
\section{Beam search}
Throughout the paper and in this supplementary material we often refer to caption results and evaluation metric results obtained using ``beam search". This section briefly summarizes our beam search procedure. While decoding the image to generate captions that describe it, rather than greedily selecting the most probable word ($N=1$), we can maintain a list of the $N$ most probable sub-sequences generated so far, generate posterior probabilities for the next word of each of these sub-sequences, and then again prune down to the $N$-best sub-sequences. This approach is widely referred to as a \emph{beam search}, where $N$ is the width of the decoding ``beam". In our experiments we additionally prune away hypotheses within the $N$-best list that have a log probability that is below that of the maximally probable partial sentence by more than $\Delta_{\log} = 5$. For all reported results, the value of $N$ is tuned on a per-model basis on the validation set (of the Karpathy splits). On MSCOCO data, $N=2$ is typically optimal for cross-entropy (XE) trained models and SCST-trained models, but in the latter case beam search provides only a very small boost in performance. For our captioning demonstrations we set $N=5$ for all models for illustrative purposes, and because we have qualitatively observed that for test images that are substantially different from those encountered during training, beam search is important.

\section{Performance of XE versus SCST  trained models}

In tables \ref{tab:SingFcVsAtt} and \ref{tab:EnsembleFcVsAtt} of the main text we compared the performance of models trained to optimize the CIDEr metric with self-critical sequence training (SCST) with that of their corresponding bootstrap models, which were trained under the cross-entropy (XE) criterion using scheduled sampling \cite{ScheduledSampling}. We provide some additional details about these experiments here. For all XE models, the probability $p_f$ of feeding forward the maximally probable word rather than the ground-truth word was increased by $0.05$ every 5 epochs until reaching a maximum value of $0.25$. The XE model with the best performance on the validation set of the Karpathy splits was then selected as the bootstrap model for SCST (with the exception of the Att2all attention models, where CE training was intentionally terminated prematurely to encourage more exploration during early epochs of RL training).

For all models, the performance of greedily decoding each word at test time is reported, as is the performance of \emph{beam search} as described in the previous section. As reported in \cite{Ranzato}, we found that beam search using RL-trained models resulted in very little performance gain. Figure \ref{fig:allmetrics} depicts the performance of our best Att2in model, which is trained to directly optimize the CIDEr metric, as a function of training epoch and evaluation metric, on the validation portion of the Karpathy splits. Optimizing CIDEr clearly improves all of the MSCOCO evaluation metrics substantially.

\section{Examples of Generated Captions}
\label{Sec:appExample}

\begin{figure*}[ht]
\vspace{-0.2in}
\begin{center}
 \includegraphics[width=0.4\linewidth]{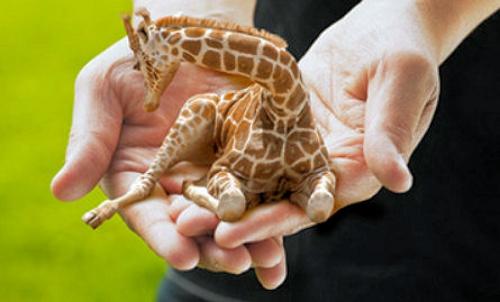}
\end{center}
   \caption{Picture of a common object in MSCOCO (a giraffe) situated in an \emph{uncommon} context (out of COCO domain) \cite{Jinchoi_contextmodels}.}
\label{fig:giraffe}
\end{figure*}

\begin{figure*}[ht]
\begin{center}
 \includegraphics[width=0.9\linewidth]{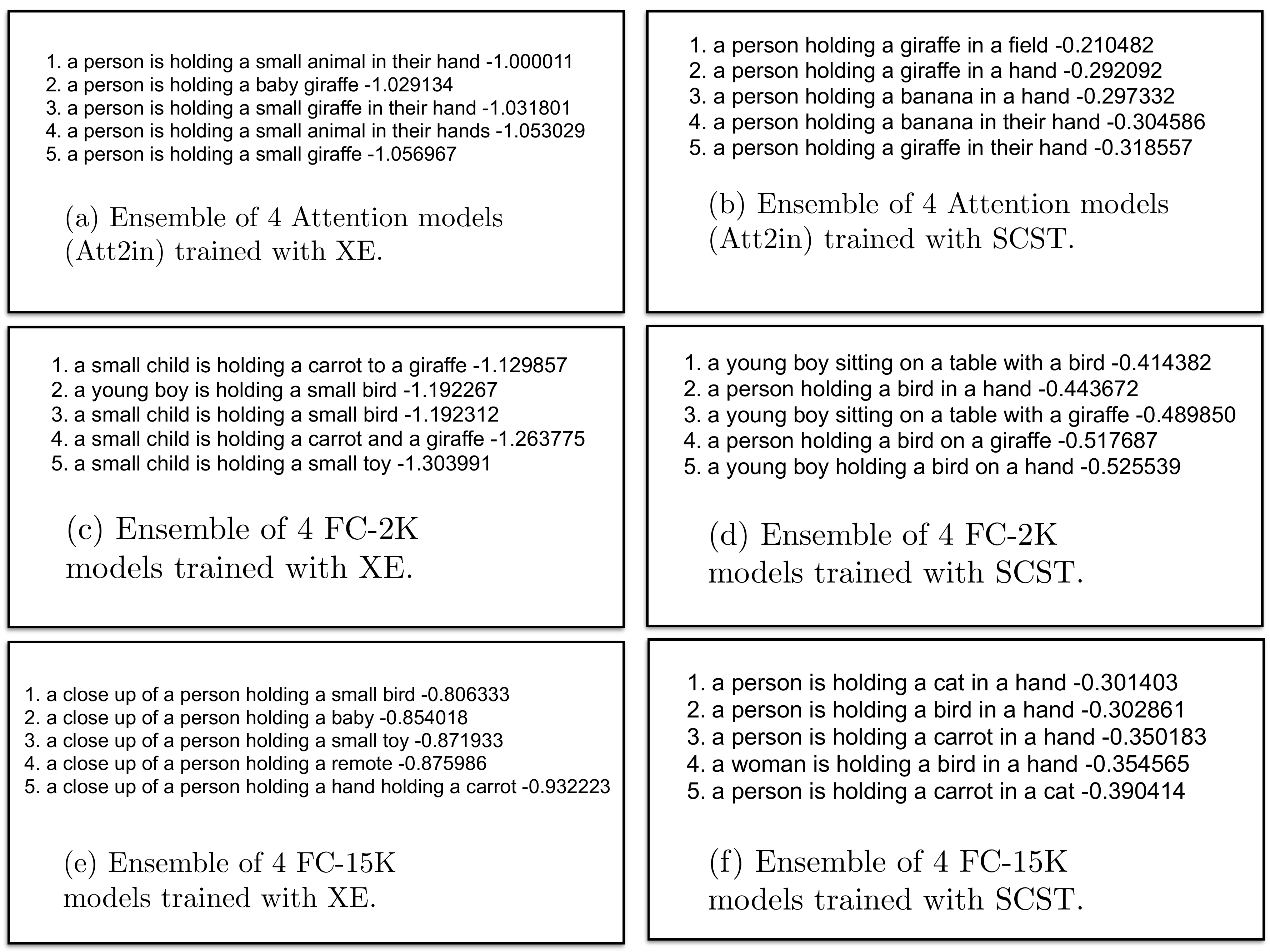}
\end{center}
   \caption{Captions generated by various models discussed in the paper to describe the image depicted in figure \ref{fig:giraffe}. Beside each caption we report the average of the log probabilities of each word, normalized by the sentence length. Notice that the attention models trained with SCST give an accurate description of this image with high confidence. Attention models trained with XE are less confident about the correct description. FC models trained with CE or SCST fail at giving an accurate description.}
\label{fig:CapGen}
\end{figure*}

\begin{figure*}[ht]
\vspace{-0.2in}
\begin{center}
 \includegraphics[width=0.4\linewidth]{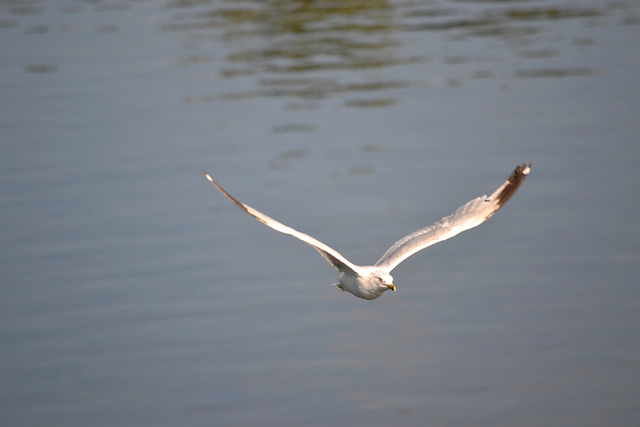}
\end{center}
   \caption{An image from the MSCOCO test set (Karpathy splits).}
\label{fig:bird}
\end{figure*}

\begin{figure*}[ht]
\begin{center}
 \includegraphics[width=0.9\linewidth]{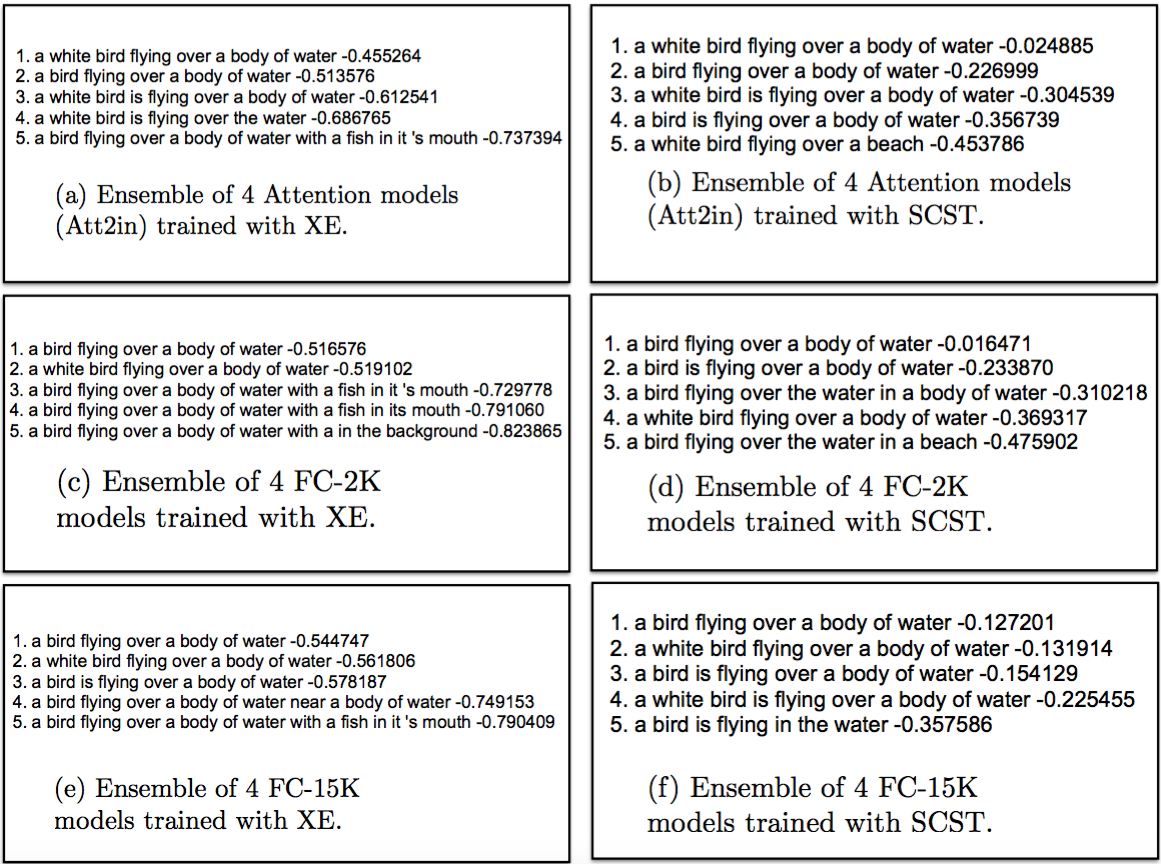}
\end{center}
   \caption{Captions generated for the image depicted in Figure \ref{fig:bird} by various models discussed in the paper. Beside each caption we report the average log probability of the words in the caption. All models perform well on this test image from the MSCOCO distribution. More generally we have observed that qualitatively, all models perform comparably on the MSCOCO test images.}
\label{fig:CapGen1}
\end{figure*}

\begin{figure*}[ht]
\vspace{-0.2in}
\begin{center}
 \includegraphics[width=0.4\linewidth]{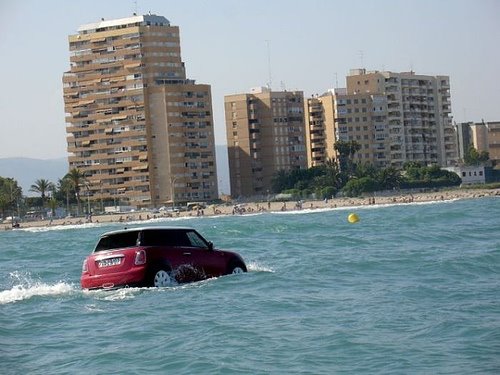}
\end{center}
   \caption{An image from the objects out-of-context (OOOC) dataset of images from \cite{Jinchoi_contextmodels}.}
\label{fig:car}
\end{figure*}

\begin{figure*}[ht]
\begin{center}
 \includegraphics[width=0.9\linewidth]{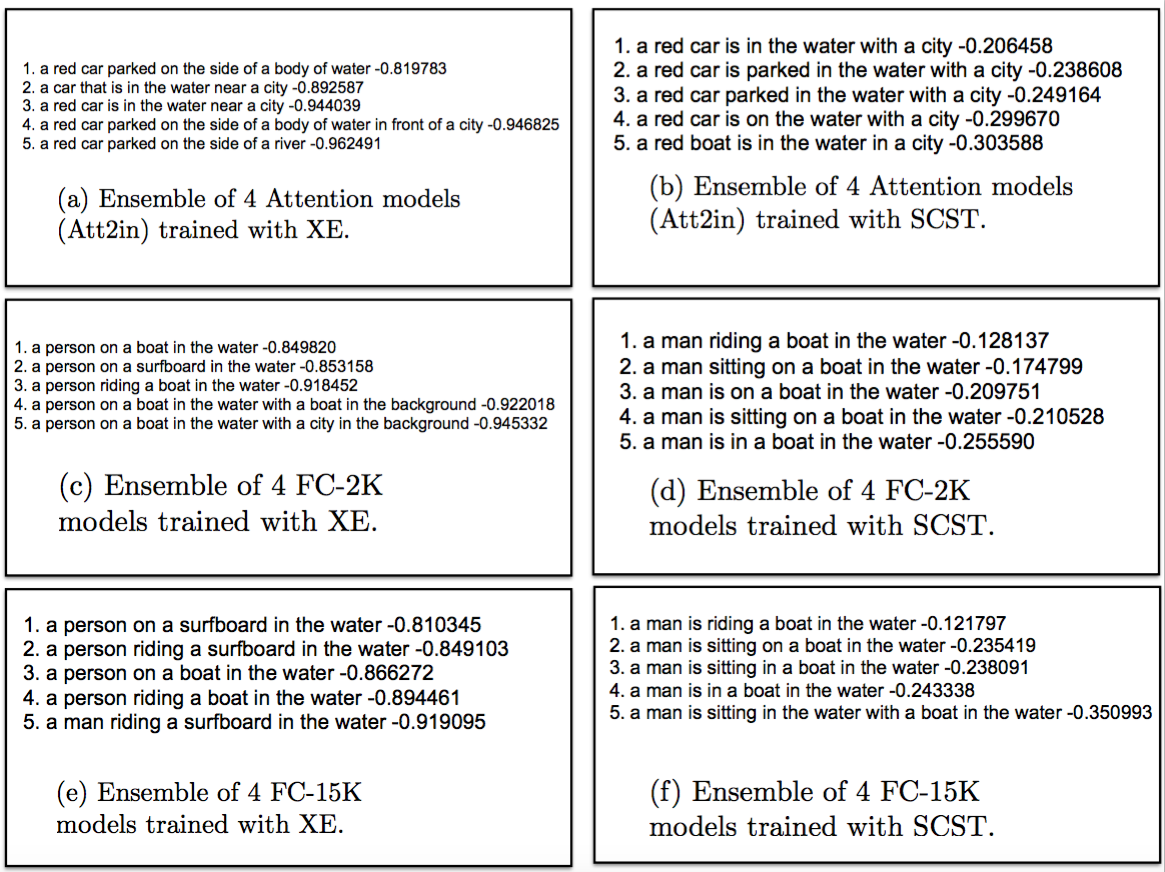}
\end{center}
   \caption{Captions generated for the image depicted in Figure \ref{fig:car} by the various models discussed in the paper. Beside each caption we report the average log probability of the words in the caption. On this image, which presents an object situated in an atypical context \cite{Jinchoi_contextmodels}, the FC models fail to give an accurate description, while the attention models handle the previously unseen image composition well. The models trained with SCST return a more accurate and more detailed summary of the image.}
\label{fig:CapGen2}
\end{figure*}

\begin{figure*}[ht]
\vspace{-0.2in}
\begin{center}
 \includegraphics[width=0.2\linewidth]{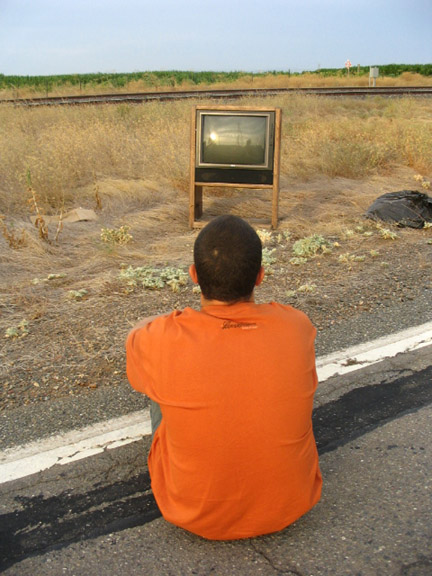}
\end{center}
   \caption{An image from the objects out-of-context (OOOC) dataset of images from \cite{Jinchoi_contextmodels}.}
\label{fig:man}
\end{figure*}

\begin{figure*}[ht]
\begin{center}
 \includegraphics[width=0.9\linewidth]{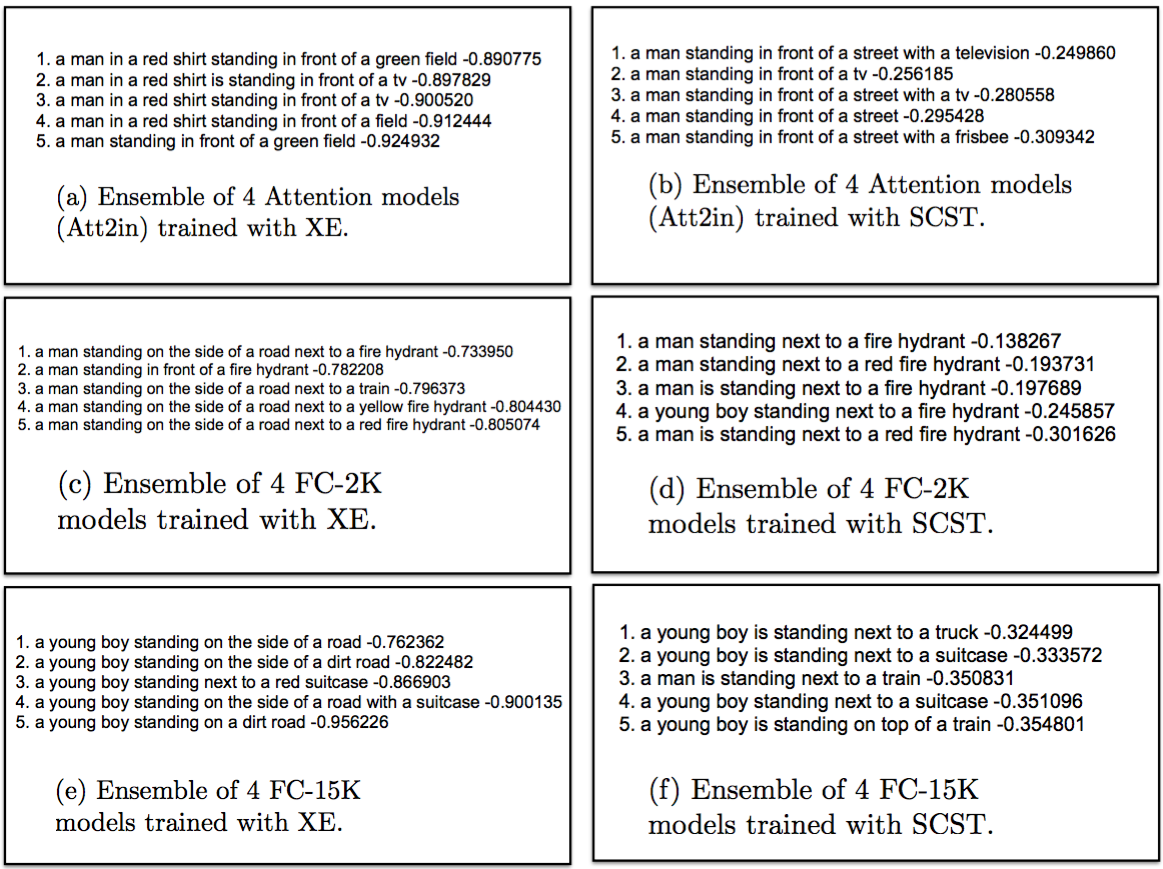}
\end{center}
   \caption{Captions generated for the image depicted in Figure \ref{fig:man} by the various models discussed in the paper. Beside each caption we report the average log probability of the words in the caption. On this image, which presents an object situated in an atypical context \cite{Jinchoi_contextmodels}, the FC models fail to give an accurate description, while the attention models handle the previously unseen image composition well. The models trained with SCST return a more accurate and more detailed summary of the image.}
\label{fig:CapGen3}
\end{figure*}

Figures \ref{fig:giraffe}-\ref{fig:CapGen3} depict demonstrations of the captioning performance of all systems. In general we found that the performance of all captioning systems on MSCOCO data is qualitatively similar, while on images containing objects situated in an uncommon context \cite{Jinchoi_contextmodels} (i.e. unlike the MSCOCO training set) our attention models perform much better, and SCST-trained attention models output yet more accurate and descriptive captions. 

\section{Further details and analysis of SCST training}

One detail that was crucial to optimizing CIDEr to produce better models was to include the EOS tag as a word. When the EOS word was omitted, trivial sentence fragments such as ``with a" and ``and a" were dominating the metric gains, despite the ``gaming" counter-measures (sentence length and precision clipping) that are included in CIDEr-D \cite{CIDEr}, which is what we optimized. Including the EOS tag substantially lowers the reward allocated to incomplete sentences, and completely resolved this issue. Another more obvious detail that is important is to associate the reward for the sentence with the first EOS encountered. Omitting the reward from the first EOS fails to reward sentence completion which leads to run-on, and rewarding any words that follow the first EOS token is inconsistent with the decoding procedure. 

\begin{figure*}
        \centering
        \begin{subfigure}[b]{0.475\textwidth}
            \centering
            \includegraphics[width=\textwidth]{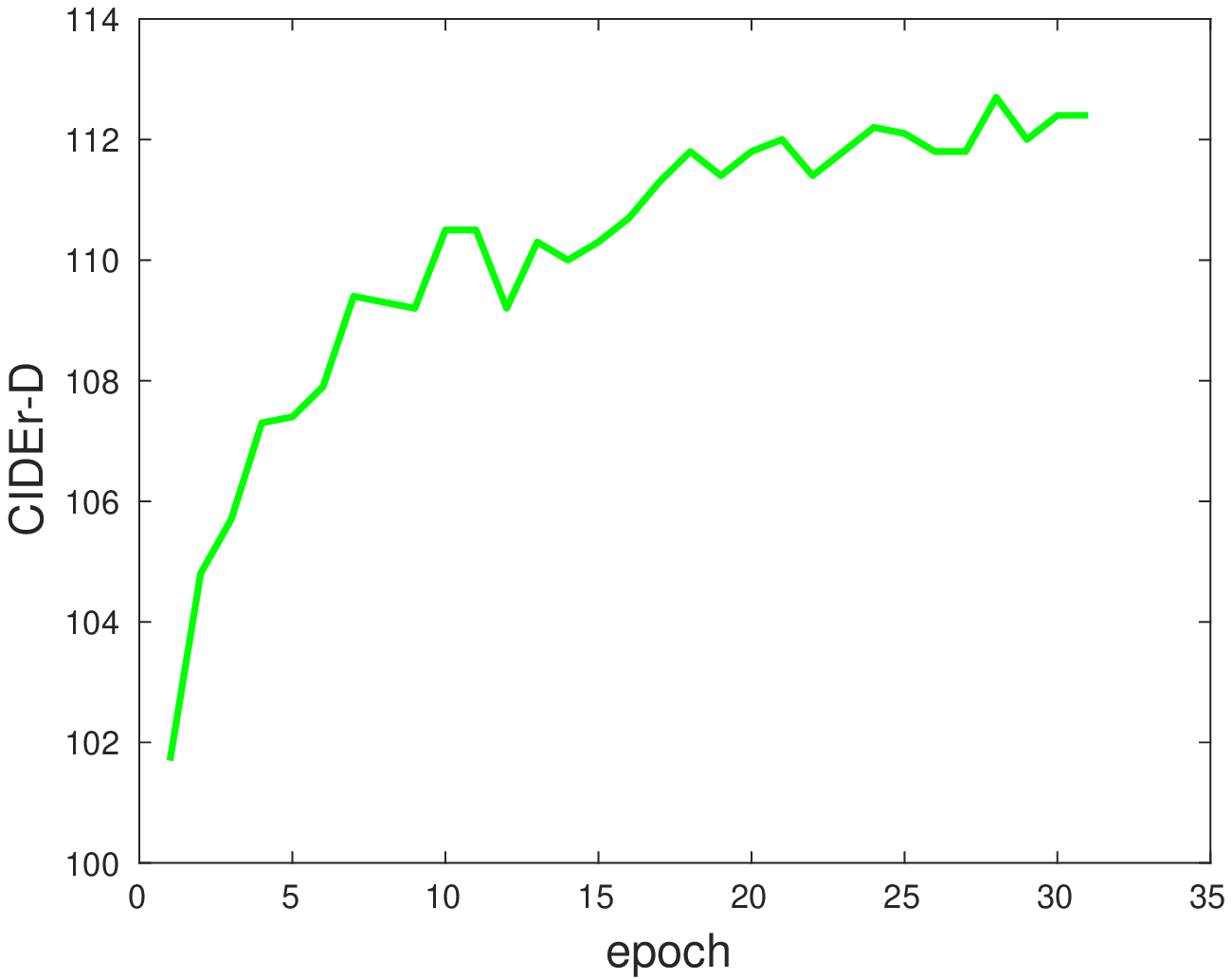}
            \label{fig:mean and std of net14}
        \end{subfigure}
        \hfill
        \begin{subfigure}[b]{0.475\textwidth}  
            \centering 
            \includegraphics[width=\textwidth]{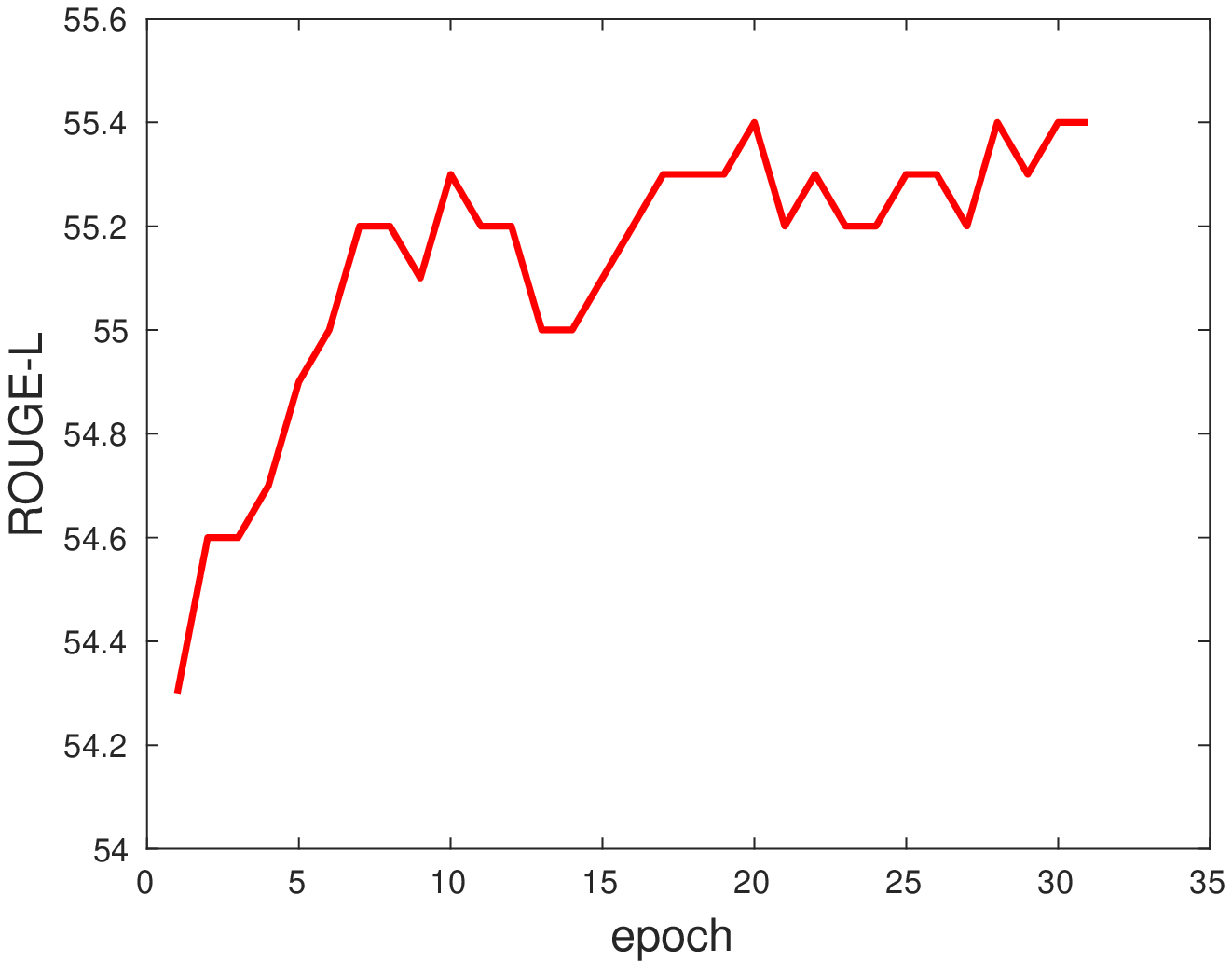}
            \label{fig:mean and std of net24}
        \end{subfigure}
        \vskip\baselineskip
        \begin{subfigure}[b]{0.475\textwidth}   
            \centering 
            \includegraphics[width=\textwidth]{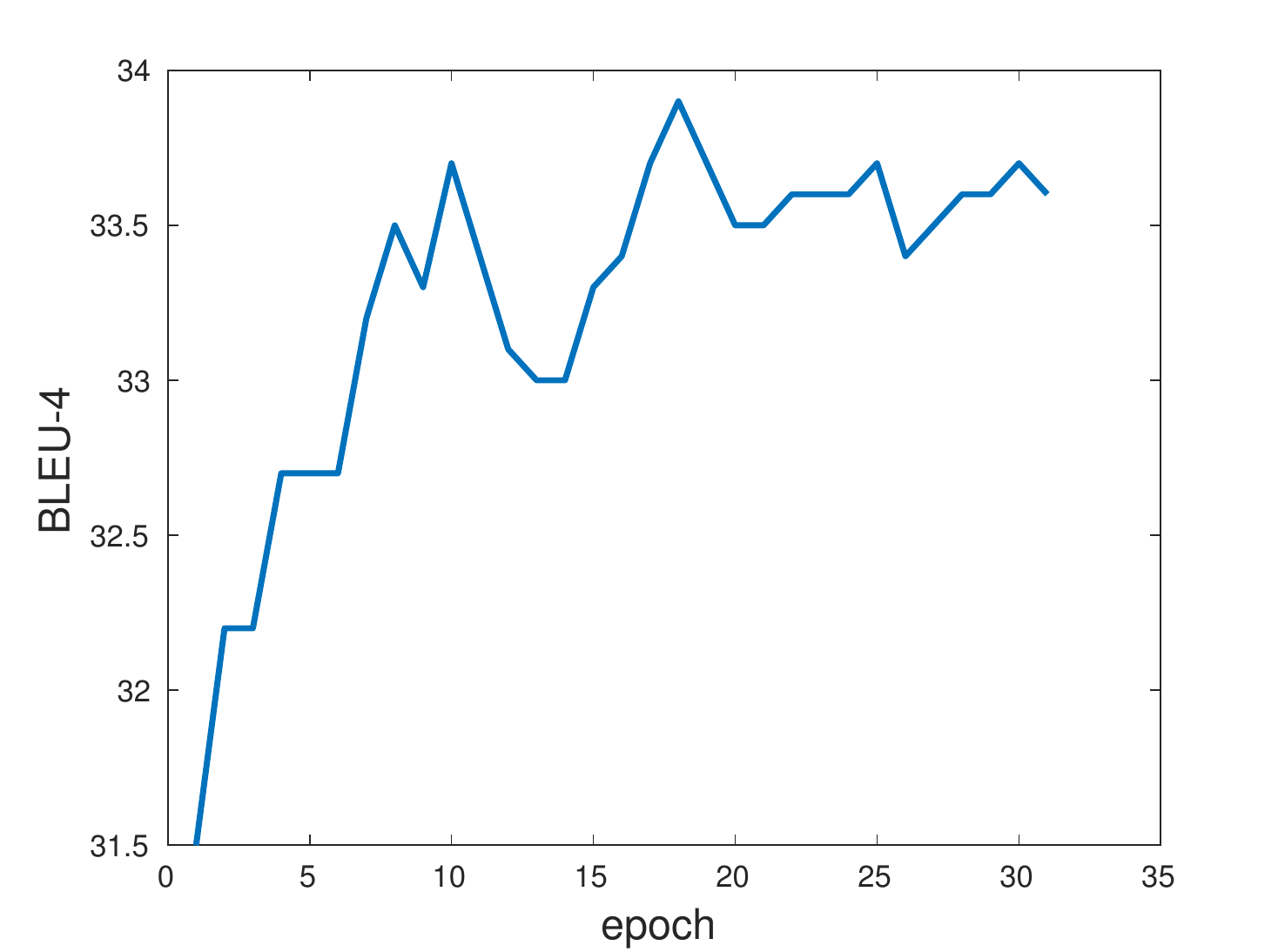}
            \label{fig:mean and std of net34}
        \end{subfigure}
        \quad
        \begin{subfigure}[b]{0.475\textwidth}   
            \centering 
            \includegraphics[width=\textwidth]{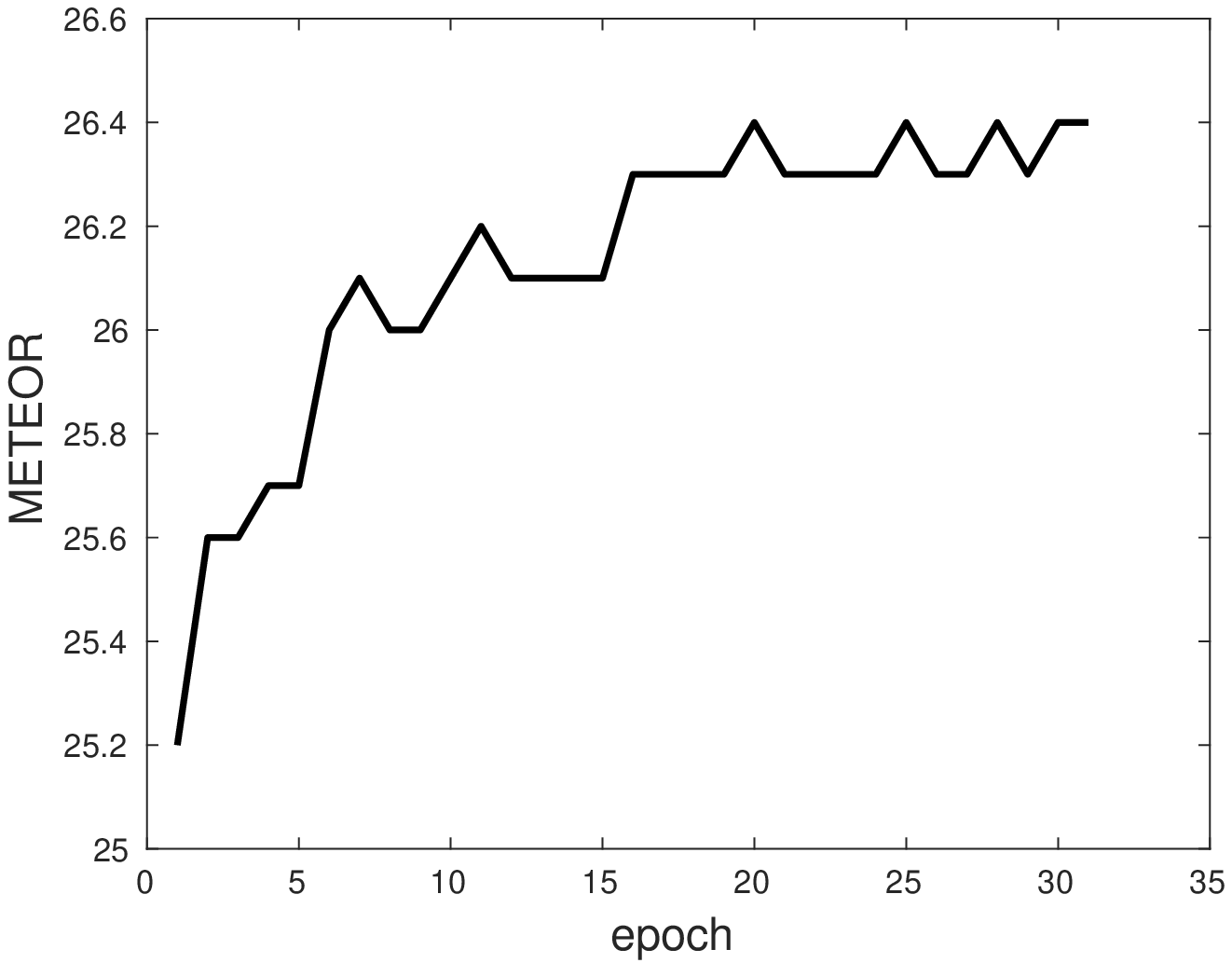}
            \label{fig:mean and std of net44}
        \end{subfigure}
        \caption[ ]%
          {Performance of our best Att2in model, which is trained to directly optimize the CIDEr metric, as a function of training epoch on the validation portion of the Karpathy splits, for the CIDEr, BLEU-4, ROUGE-L, and METEOR MSCOCO evaluation metrics. Optimizing CIDEr improves all of these evaluation metrics substantially.} 
       \label{fig:allmetrics}
    \end{figure*}

This work has focused on optimizing the CIDEr metric, since, as discussed in the paper, optimizing CIDER substantially improves all MSCOCO evaluation metrics, as was shown in tables \ref{tab:SingFcVsAtt} and \ref{tab:EnsembleFcVsAtt} and is depicted in figure \ref{fig:allmetrics}. Nevertheless, directly optimizing another metric does lead to higher evaluation scores on that same metric as shown, and so we have started to experiment with including models trained on Bleu, Rouge-L, and METEOR in our Att2in ensemble to attempt to improve it further. So far we have not been able to substantially improve performance w.r.t. the other metrics without more substantially degrading CIDEr. 

\vskip -0.2in

\clearpage

\end{document}